\documentclass[10pt,twocolumn,letterpaper]{article}

\usepackage{cvpr}              
\usepackage{makecell}
\usepackage{booktabs}
\usepackage{tabularx}
\usepackage{multirow}
\usepackage[accsupp]{axessibility} 

\newlength{\itemheight} %

\usepackage[dvipsnames]{xcolor}

\definecolor{cvprblue}{rgb}{0.21,0.49,0.74}
\usepackage[pagebackref,breaklinks,colorlinks,citecolor=cvprblue]{hyperref}

\def\paperID{5939} 
\def\confName{CVPR}
\def\confYear{2024}

\title{DyBluRF: Dynamic Neural Radiance Fields from Blurry Monocular Video}

\author{Huiqiang Sun$^{1}$\hspace{0.1in} 
        Xingyi Li$^{1}$\hspace{0.1in} 
        Liao Shen$^{1}$\hspace{0.1in} 
        Xinyi Ye$^{1}$\hspace{0.1in} 
        Ke Xian$^{2}$\hspace{0.1in}
        Zhiguo Cao$^{1}$\footnotemark[1]~\hspace{0.1in} \\
$^1$School of AIA, Huazhong University of Science and Technology\\
$^2$School of EIC, Huazhong University of Science and Technology\hspace{0.3in}\\
{\tt\small \{shq1031,xingyi\_li,leoshen,xinyiye,kxian,zgcao\}@hust.edu.cn}\\
{\small{\url{https://huiqiang-sun.github.io/dyblurf}}}
\vspace{-2mm}
}

\begin{document}

\twocolumn[{%
\renewcommand\twocolumn[1][]{#1}%
\maketitle
\centering
\setlength{\itemheight}{2.97cm}

\includegraphics[width=\textwidth]{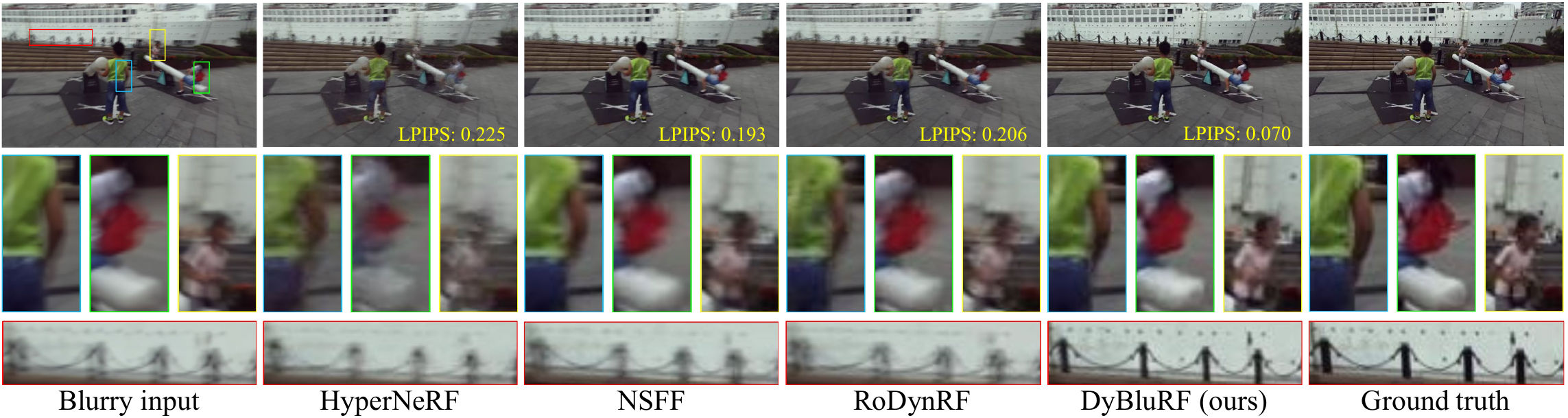}
\vspace{-0.6cm}
{\captionof{figure}{\label{fig:fig1} 
Given a monocular video capturing a dynamic scene with motion blur, our proposed method, DyBluRF, effectively synthesizes high-quality and sharp novel views compared to previous dynamic NeRF approaches that often yield low-quality and blurry results.
}}
\vspace{2.0em}
}]
\renewcommand{\thefootnote}{\fnsymbol{footnote}} 
\footnotetext[1]{Corresponding author.}

\begin{abstract}
Recent advancements in dynamic neural radiance field methods have yielded remarkable outcomes. However, these approaches rely on the assumption of sharp input images. When faced with motion blur, existing dynamic NeRF methods often struggle to generate high-quality novel views. In this paper, we propose DyBluRF, a dynamic radiance field approach that synthesizes sharp novel views from a monocular video affected by motion blur. To account for motion blur in input images, we simultaneously capture the camera trajectory and object Discrete Cosine Transform (DCT) trajectories within the scene. Additionally, we employ a global cross-time rendering approach to ensure consistent temporal coherence across the entire scene. We curate a dataset comprising diverse dynamic scenes that are specifically tailored for our task. Experimental results on our dataset demonstrate that our method outperforms existing approaches in generating sharp novel views from motion-blurred inputs while maintaining spatial-temporal consistency of the scene.
\end{abstract}    
\section{Introduction}
\label{sec:intro}

The research on novel view synthesis for dynamic scenes has gained substantial importance in the community, with practical applications in a wide range of AR/VR contexts. Recent breakthroughs can be attributed to Neural Radiance Fields (NeRFs)~\cite{mildenhall2020nerf}. NeRFs have made a profound impact on novel view synthesis, and recent years have witnessed rapid advancements in NeRF, including several works dedicated to dynamic scene representation~\cite{pumarola2021d, park2021hypernerf, li2021neural, gao2021dynamic, liu2023robust}. Despite producing high-quality novel views, these dynamic NeRF methods may exhibit significant performance degradation when presented with inputs containing motion blur, as shown in Fig.~\ref{fig:fig1}. Recently, some NeRF-based methods have emerged to tackle the motion deblurring of static scenes~\cite{ma2022deblur, lee2023dp, wang2023bad, lee2023exblurf}. 
However, these methods cannot address blurring caused by object motion and lack the capability to effectively represent dynamic scenes.

To the best of our knowledge, 
there is no NeRF-based method dedicated to dynamic scene deblurring. Yet, in real-life video captures of dynamic scenarios, extended camera exposure time often introduces motion blur. Consequently, obtaining a sharp dynamic radiance field from inputs affected by motion blur remains a pressing problem. Existing dynamic NeRF methods heavily rely on sharp inputs to accurately represent dynamic scenes. When motion blur is present, the modulation of inter-frame relationships becomes unreliable, leading to difficulties in representing object motion and maintaining temporal consistency in dynamic scenes. 
Moreover, the presence of blur in input data degrades certain data priors like depth and optical flow, leading to inaccurate constraining of scene geometry. 
These challenges significantly hinder the construction of sharp dynamic NeRF representations from blurry inputs. 

In this work, we propose DyBluRF, a model specifically designed to take monocular videos of dynamic scenes with motion blur as input and generate sharp novel views. Motion blur in dynamic scenes typically results from both camera and object movements. To model camera motion, drawing inspiration from BAD-NeRF~\cite{wang2023bad}, we evenly discretize the exposure time into multiple timestamps and learn camera poses at these timestamps. For object motion representation, establishing inter-frame relationships, such as directly predicting $3$D scene flow, is unreliable under conditions of blurry input, thus failing to accurately present dynamic scenes. Instead, we employ an MLP to predict the global DCT trajectories of scene objects~\cite{valmadre2012general}, thereby simulating object motion blur by expressing the DCT trajectories within an exposure time. 

To enhance the temporal consistency representation in dynamic scenes, we adopt a cross-time rendering approach based on the input blurry images. This method uses the predicted DCT trajectory to model scene correlation across multiple input views and employs deformed rays for rendering by integrating scene information from other frames into the target frame. To ensure a globally consistent temporal expression, cross-time rendering is not limited to adjacent frames but instead performed across the entire temporal range. At the initial stages of training, frames close to the target frame are selected for cross-time rendering. As the training progresses, frames further away from the target frame are included until encompass the entire temporal range. This cross-time rendering approach further strengthens the representation of temporal consistency. 

Furthermore, it is not reasonable to directly use depth and optical flow maps generated from blurry images to constrain the model, as the predictions of depth and optical flow can become inaccurate due to blurriness. Therefore, we introduce data-driven priors to mitigate the impact of inaccuracies in depth and optical flow from blurry images. 

Given the absence of existing dynamic scene blurring datasets based on NeRF, we curate a collection of dynamic scenes sourced from the Stereo Blur Dataset~\cite{zhou2019davanet}. We use these scenes for comparative analysis of our approach against existing dynamic NeRF methods, demonstrating its effective handling of input images with motion blur and generation of high-quality novel views. We also preprocess the input blurry images using state-of-the-art single-image deblurring and dynamic scene video deblurring methods before feeding them into existing dynamic NeRF methods. Experimental results demonstrate that our method still outperforms existing dynamic NeRF methods. Additionally, we conduct ablation studies to validate the effectiveness of each component of our approach. In summary, our key contributions include:

\begin{itemize}[leftmargin=0.6cm]
    \item We introduce DyBluRF, the first dynamic NeRF model specifically designed to effectively tackle motion-blurred monocular video.
    \item We propose a dynamic scene representation method based on DCT, simulating the motion blur induced by object movement along the DCT trajectory. 
    \item We present a cross-time rendering method that ensures temporal consistency by modeling scene relationships across different exposure times.
    \item We propose an effective approach that incorporates depth maps and optical flow derived from blurry images to impose constraints on NeRF.
\end{itemize}
\section{Related Work}
\label{sec:related work}

\noindent\textbf{Image Deblurring.}
The main goal of image deblurring is to restore a sharp image from a blurry one. 
From the perspective of input data types, current motion deblurring methods can be categorized into single-image deblurring and multi-image/video deblurring. In the case of single-image deblurring, 
Some traditional methods employ unified probabilistic model~\cite{shan2008high}, normalized gradient sparsity~\cite{krishnan2011blind}, or model rotational velocity of the camera during exposure time~\cite{whyte10} to mitigate motion blur. Recently, numerous deep learning-based image deblurring methods have emerged~\cite{nah2017deep, kupyn2018deblurgan, tao2018scale, kupyn2019deblurgan, zamir2022restormer}, accompanied by the introduction of several datasets dedicated to image deblurring~\cite{shen2019human}. These methods often achieve superior results compared to traditional techniques.

In contrast to single-image deblurring, video deblurring requires establishing temporal consistency among different frames to ensure temporal smoothness. Some approaches employ optical flow information to establish inter-frame relationships~\cite{hyun2015generalized, pan2020cascaded}. However, the optical flow in blurry videos often lacks accuracy. 
With the advance of deep learning, many techniques based on CNNs~\cite{su2017deep}, RNNs~\cite{zhong2020efficient}, or attention mechanisms~\cite{zhang2022spatio} have been proposed to address the problem of video deblurring. While these methods yield excellent results, they predominantly operate in image space and cannot capture scene geometry information in $3$D space. Our approach leverages NeRF to represent $3$D information of dynamic scenes implicitly and establishes spatial-temporal consistency to achieve deblurring. 

\noindent\textbf{Neural Radiance Field.}
NeRF~\cite{mildenhall2020nerf} has demonstrated impressive results in novel view synthesis, and subsequent works have aimed to enhance it from various perspectives, including fundamental enhancements~\cite{barron2021mip, barron2022mip}, fast training and rendering~\cite{muller2022instant, chen2022tensorf, fridovich2022plenoxels, chen2023mobilenerf}, pose estimation~\cite{wang2021nerf, lin2021barf, bian2023nope}, or sparse viewpoint input~\cite{yu2021pixelnerf, yang2023freenerf}. Since vanilla NeRF imposes high requirements on input image quality, several follow-up methods have explored generating high-quality novel views in degraded input conditions, such as in-the-wild images~\cite{martin2021nerf}, input images with burst noise~\cite{pearl2022nan}, or low dynamic range (LDR) inputs with varying exposures~\cite{huang2022hdr}. Recently, several NeRF-based approaches have been devised to handle blurry inputs. Deblur-NeRF~\cite{ma2022deblur} and DP-NeRF~\cite{lee2023dp} restore sharp radiance fields by modeling a blur kernel from various blur types. BAD-NeRF~\cite{wang2023bad}, which is most relevant to our work, models motion blur by estimating camera trajectory. ExBluRF~\cite{lee2023exblurf} leverages a voxel-based radiance field to handle extreme motion blur. While these methods can remove blur from input images and generate high-quality novel views, they cannot handle dynamic scenes. Our approach can effectively reduce motion blur caused by both camera and object motion, generating novel views of dynamic scenes.

\noindent\textbf{Dynamic Neural Radiance Field.}
Many recent efforts have extended NeRF to represent dynamic scenes. These methods input multi-view videos~\cite{li2022neural} or a monocular video~\cite{xian2021space, li2023dynibar} of dynamic scenes to achieve space-time novel view synthesis. To ensure temporal consistency, some approaches represent dynamic scenes by deforming the canonical space points~\cite{pumarola2021d, tretschk2021non, park2021nerfies, park2021hypernerf, wang2023flow}. Other methods leverage optical flow constraints to establish scene coherence between adjacent frames~\cite{li2021neural, gao2021dynamic, wang2021neural, liu2023robust}. In particular, NSFF~\cite{li2021neural} employs neural scene flow fields to represent complex object motions in dynamic scenes; RoDynRF~\cite{liu2023robust} jointly learns camera parameters to enhance model robustness. Additionally, many methods are currently dedicated to fast training and rendering of dynamic NeRF~\cite{fang2022fast, cao2023hexplane, attal2023hyperreel}. However, these dynamic NeRF methods require sharp input images and often struggle to synthesize reliable novel views in the presence of motion blur in the input images.
\section{Method}
\label{sec:method}


\begin{figure*}
    \centering
    \includegraphics[width=1.0\linewidth]{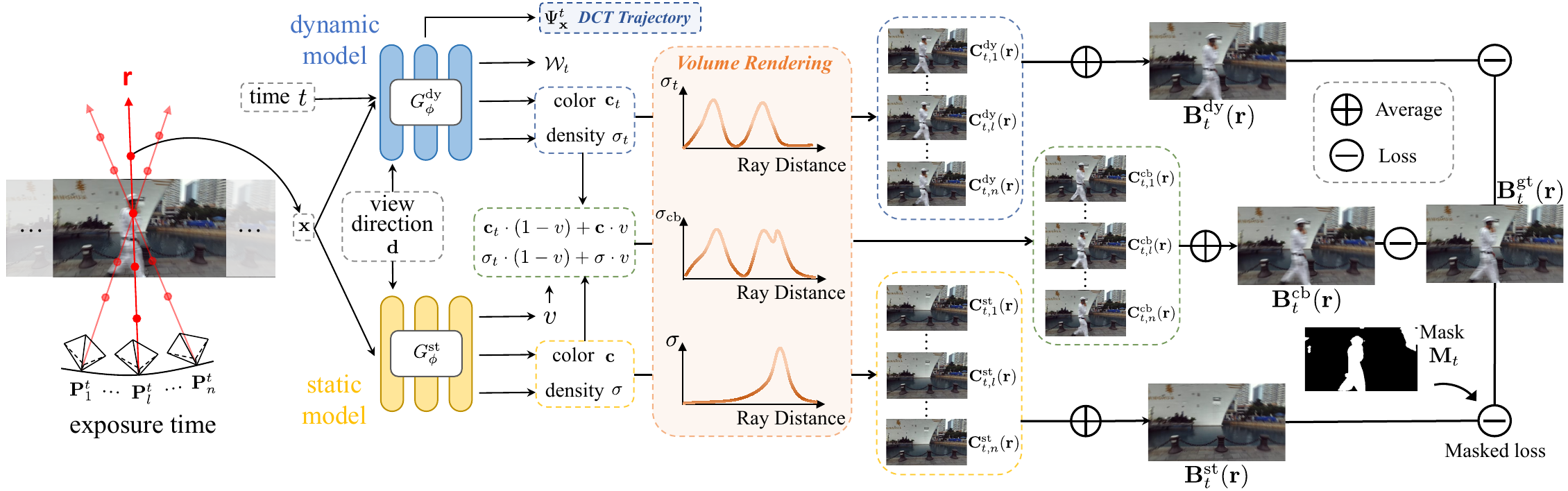}
    \caption{\textbf{Overall pipeline.} To model the motion blur of input images, we discretize the exposure time into $n$ timestamps. Subsequently, we perform ray sampling for the same pixel based on the camera poses for each timestamp. We employ two MLPs to represent dynamic scenes. The dynamic model takes spatial point coordinates $\mathbf{x}$, viewing direction $\mathbf{d}$, and time $t$ as inputs and predicts color $\mathbf{c}_t$, volume density $\sigma_t$, DCT coefficients $\Psi_{\mathbf{x}}^t$, and disocclusion weights $\mathcal{W}_t$. The static model only takes $\mathbf{x}$ and $\mathbf{d}$ as inputs and predicts color $\mathbf{c}$, volume density $\sigma$, and a blending weight $v$ for blending static and dynamic results. After obtaining colors and volume densities for static, dynamic, and blended scenes, we use volume rendering to compute pixel RGB values $\mathbf{C}(\mathbf{r})$. We average the RGB values for the same pixel within the exposure time to obtain the predicted blurry image, and calculate losses with the input blurry image. For dynamic and blended images, losses are directly computed against the input image. For the static loss, we use a mask image to calculate the static regions only.}
    \label{fig:pipeline}
\end{figure*}

We propose DyBluRF, a dynamic radiance field approach synthesizing sharp novel views from a monocular video affected by motion blur. The overall pipeline is illustrated in Fig.~\ref{fig:pipeline}. In this section, we first provide background information on neural radiance fields and volume rendering in Section~\ref{sec:method_1}. We then elaborate on modeling genuine physical image formation process of motion blur in Section~\ref{sec:method_2}. Next, Section~\ref{sec:method_3} outlines our approach for representing dynamic scenes and ensuring temporal consistency within the scene. Finally, we provide details of the training and optimization process in Section~\ref{sec:method_4}.

\subsection{Preliminary: Neural Radiance Field}
\label{sec:method_1}
We begin by describing Neural Radiance Fields (NeRFs) that our method builds upon. NeRF represents a scene as an implicit function that inputs a spatial point location $\mathbf{x} \in \mathbb{R}^3$ and viewing direction $\mathbf{d} \in \mathbb{R}^3$, and outputs corresponding volume density $\sigma$ and color $\mathbf{c}$:
\begin{equation}
    (\mathbf{c}, \sigma) = F_{\Theta}(\mathbf{x}, \mathbf{d})\,.
\end{equation}
To enable learning scene details, point location $\mathbf{x}$ and viewing direction $\mathbf{d}$ are transformed into a high-dimensional space through positional embedding $\gamma(\cdot)$ before sending to NeRF. We also employ this trick in our methods but omit to mention the position encoding in our writing for clarity.

NeRF adopts classical volume rendering~\cite{kajiya1984ray} to render RGB images from any given camera pose. For a ray $\mathbf{r}$ emitted from camera center through a given pixel 
on the image plane, 
its color is computed as: 
\begin{equation}
\label{eq:volume rendering}
    \mathbf{C}(\mathbf{r}) = \int_{s_n}^{s_f} T(s) \thinspace \sigma (\mathbf{r}(s)) \thinspace \mathbf{c}(\mathbf{r}(s), \mathbf{d}) ds\,,
\end{equation}
where
\begin{equation}
    T(s) = \exp{\left(-\int_{s_n}^s \sigma (\mathbf{r}(m)) dm\right)}\, 
\end{equation}
is accumulated transmittance along ray $\mathbf{r}$ which denotes the probability that the ray will not hit any object from $s_n$ to $s$.

\subsection{Motion Blur Formulation}
\label{sec:method_2}
The physical process of motion blur can be considered as the result of camera or object motion within exposure time $\tau$. The mathematical modeling of the motion blur generation process can be written as:
\begin{equation}
    \mathbf{B}(\mathrm{x}) = g \int_0^{\tau} \mathbf{C}_t(\mathrm{x}) dt\, ,
\end{equation}
where $\mathbf{C}_t(\mathrm{x})$ denotes the sharp image captured at time $t$, $\mathbf{B}(\mathrm{x})$ is the blurry image, $g$ is a normalization factor. To approximate the integral operation, we discretize the exposure time into $n$ timestamps uniformly and consider the captured blurry image as the average of $n$ sharp images during the exposure time: 
\begin{equation}
\label{eq:blur_process}
    \mathbf{B}(\mathrm{x}) \approx \frac{1}{n} \sum_{l=1}^{n} \mathbf{C}_l(\mathrm{x})\, .
\end{equation}
In theory, as $n$ increases, the simulated blur generation process becomes more close to a genuine one.

According to Eq.~\ref{eq:blur_process}, generating a blurry image demands obtaining sharp images at each timestamp within the exposure time. Therefore, it becomes crucial to model both camera and object motion during the exposure time. For the object motion representation, we will provide a detailed explanation in Section~\ref{sec:method_3}. For the camera motion, we adopt a strategy that jointly learns camera parameters and NeRF. Specifically, given $N$ input blurry images, we initialize the camera poses $\mathbf{P}_1^i$ and $\mathbf{P}_n^i$ at the beginning and end timestamps of exposure time $i$, where $i \in \{1, \cdots, N\}$. Leveraging the short exposure time, we utilize linear interpolation to obtain camera poses for intermediate timestamps. Following the ideas of BAD-NeRF~\cite{wang2023bad}, we employ Lie-algebra of $SE(3)$ to interpolate the intermediate camera poses $\mathbf{P}_l^i$, $l \in \{2,\cdots,n-1\}$.

\subsection{Temporal Consistency Modeling}
\label{sec:method_3}
\noindent\textbf{Dynamic radiance field.}
Similar to existing dynamic NeRF methods, we employ an MLP to represent dynamic scenes. This MLP takes the input spatial point location $\mathbf{x}$, viewing direction $\mathbf{d}$, and time $t$, and outputs the corresponding color and volume density: $G_{\phi}^{\textrm{dy}}: (\mathbf{c}_t, \sigma_t) = G_{\phi}^{\textrm{dy}}(\mathbf{x}, \mathbf{d}, t)$, where $G_{\phi}^{\textrm{dy}}$ indicates the MLP parameterized by $\phi$. Handling a monocular video where scene information at different times is limited to individual frames poses an ill-posed problem. Establishing temporal consistency in dynamic scenes becomes essential to resolve this issue. Although employing scene flow information between adjacent frames is a direct approach~\cite{li2021neural, liu2023robust}, accurately determining scene motion within the exposure time using adjacent blurry images remains challenging. To enforce a more reliable temporal consistency, we establish relationships across global time intervals rather than relying solely on adjacent frames. Leveraging the efficiency of the Discrete Cosine Transform (DCT) in representing complex motion trajectories in a smooth and concise manner~\cite{valmadre2012general, wang2021neural}, we choose to model the object DCT trajectory to express dynamic scenes.

Given a sequence of $N$ blurry images, the exposure time for each image can be divided into $n$ timestamps, resulting in a total of $n \times N$ timestamps. The trajectory with DCT coefficients $\Psi \in \mathbb{R}^{3K}$ can be expressed as: 
\begin{equation}
\label{eq:DCT}
    \mathcal{T}(t) = \sqrt{\frac{2}{nN}} \sum_{k=1}^K \Psi(k) \cos \left( \frac{\pi}{2nN} (2t + 1)k \right)\, ,
\end{equation}
where $\mathcal{T} \in \mathbb{R}^3$ represents the time-dependent trajectory, $\Psi(k) \in \mathbb{R}^3$ denotes the $k$-th coefficient in $3$D space, $K$ is a hyperparameter. According to Eq.~\ref{eq:DCT}, for a $3$D point $\mathbf{x}$ at time $t$, its global time motion trajectory $\mathcal{T}_{\mathbf{x}}^t$ can be modeled using the corresponding DCT coefficient $\Psi_{\mathbf{x}}^t$, and $\mathcal{T}_{\mathbf{x}}^t(t')$ represents its $3$D coordinates at time $t'$. We use the MLP to predict the DCT coefficients. Thus, the dynamic radiance field can be rewritten as $G_{\phi}^{\textrm{dy}}: (\mathbf{c}_t, \sigma_t, \Psi_{\mathbf{x}}^t) = G_{\phi}^{\textrm{dy}}(\mathbf{x}, \mathbf{d}, t)$. 

To handle occlusion in dynamic scenes, we adopt a similar approach to \cite{li2021neural}, where the MLP additionally outputs a disocclusion weight $\mathcal{W}_t = (w_{\textrm{fw}}, w_{\textrm{bw}})$. However, unlike \cite{li2021neural}, our method predicts disocclusion weights to represent the occlusion possibility to the corresponding timestamps within the subsequent or preceding exposure time. In summary, our dynamic radiance fields $G_{\phi}^{\textrm{dy}}$ can be expressed as:
\begin{equation}
\label{eq:MLP}
    (\mathbf{c}_t, \sigma_t, \Psi_{\mathbf{x}}^t, \mathcal{W}_t) = G_{\phi}^{\textrm{dy}}(\mathbf{x}, \mathbf{d}, t)\, .
\end{equation}

To supervise the model using input images, we need to obtain the blurry image at each frame through the model. For the $i$-th blurry image $\mathbf{B}_i(\mathbf{r})$, we calculate the timestamps $t^i_l$, $l \in \{1, \cdots, n\}$ within the $i$-th camera exposure time. As per Section~\ref{sec:method_2}, the camera pose for each timestamp is denoted as $\mathbf{P}_l^i$. Using the MLP depicted in Eq.~\ref{eq:MLP} and the volume rendering from Eq.~\ref{eq:volume rendering}, we derive predicted sharp images $\mathbf{C}_{i,l}^{\textrm{dy}}(\mathbf{r})$ of timestamps $l$ within the $i$-th exposure time. Then, we average these sharp images to obtain the predicted blurry image $\mathbf{B}_i^{\textrm{dy}}(\mathbf{r})$: 
\begin{equation}
    \mathbf{B}_i^{\textrm{dy}}(\mathbf{r}) = \frac{1}{n} \sum_{l=1}^{n} \mathbf{C}_{i,l}^{\textrm{dy}}(\mathbf{r}), \ i \in \{1, \cdots, N\}\, ,
\end{equation}
and calculate rendering loss with input blurry image $\mathbf{B}^{\textrm{gt}}_i(\mathbf{r})$: 
\begin{equation}
\label{eq:dy_rgb_loss}
    \mathcal{L}_{\textrm{RGB}}^{\textrm{dy}} = \Vert \mathbf{B}_i^{\textrm{dy}}(\mathbf{r}) - \mathbf{B}^{\textrm{gt}}_i(\mathbf{r}) \Vert_2^2\, .
\end{equation}

\noindent\textbf{Static model integration.}
Similar to many previous dynamic NeRF methods~\cite{li2021neural, liu2023robust}, we also employ an additional MLP to enhance the representation of static regions: 
\begin{equation}
    (\mathbf{c}, \sigma, v) = G_{\phi}^{\textrm{st}}(\mathbf{x}, \mathbf{d})\, ,
\end{equation}
where $v$ denotes the blending weight used to combine the results from static and dynamic models: 
\begin{equation}
    \begin{aligned}
        \mathbf{c}_{\textrm{combine}} &= \mathbf{c}_t \cdot (1 - v) + \mathbf{c} \cdot v\, , \\
        \sigma_{\textrm{combine}} &= \sigma_t \cdot (1 - v) + \sigma \cdot v\, .
    \end{aligned}
\end{equation}
We utilize Eq.~\ref{eq:volume rendering} to obtain static sharp images $\mathbf{C}_{i,l}^{\textrm{st}}(\mathbf{r})$ and combined sharp images $\mathbf{C}_{i,l}^{\textrm{cb}}(\mathbf{r})$, and we average these images in an exposure time to obtain the predicted static blurry image $\mathbf{B}_i^{\textrm{st}}(\mathbf{r})$ and combined blurry image $\mathbf{B}_i^{\textrm{cb}}(\mathbf{r})$. Subsequently, we apply rendering loss from both static and combined results to constrain the MLPs: 
\begin{equation}
    \begin{aligned}
        \mathcal{L}_{\textrm{RGB}}^{\textrm{cb}} &= \Vert \mathbf{B}_i^{\textrm{cb}}(\mathbf{r}) - \mathbf{B}^{\textrm{gt}}_i(\mathbf{r}) \Vert_2^2\, , \\
        \mathcal{L}_{\textrm{RGB}}^{\textrm{st}} = (1 &- \mathbf{M}_i) \cdot \Vert \mathbf{B}_i^{\textrm{st}}(\mathbf{r}) - \mathbf{B}^{\textrm{gt}}_i(\mathbf{r}) \Vert_2^2\, .
    \end{aligned}
\end{equation}
where $\mathbf{M}_i$ denotes the motion mask of input frame $i$. The inclusion of static model enhances the ability to represent static scenes.

\begin{figure}[t]
    \centering
    \includegraphics[width=0.93\linewidth]{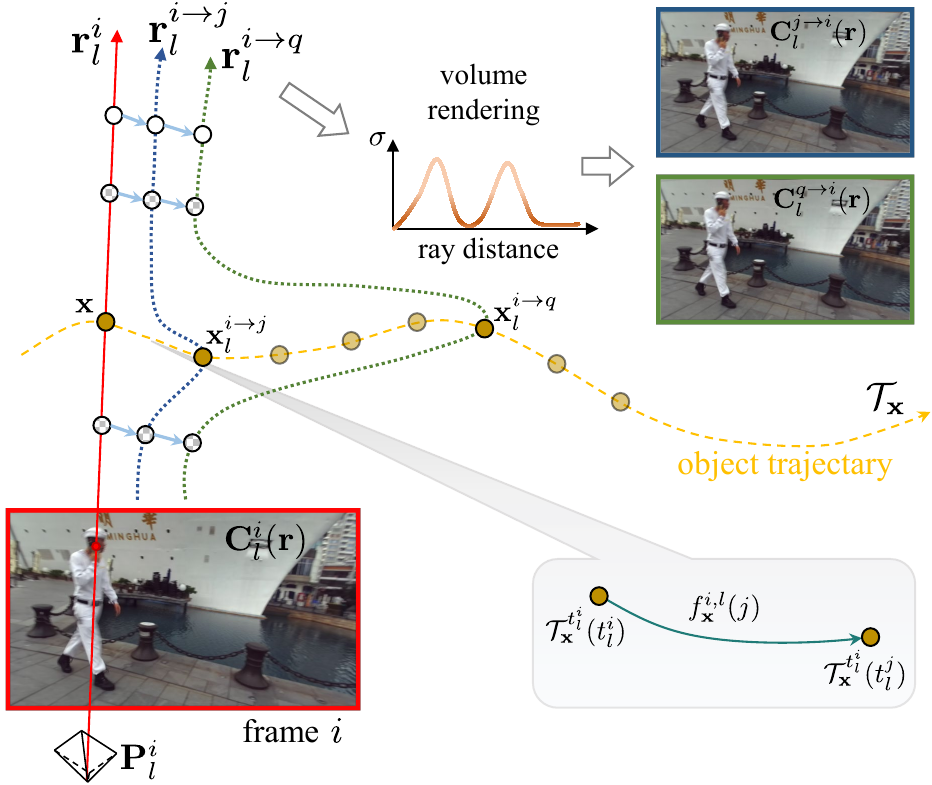}
    \caption{\textbf{Cross-time rendering.} We use cross-time rendering to ensure temporal consistency in dynamic scene representation. We render the target timestamp image $\mathbf{C}_l^i(\mathbf{r})$ in frame $i$ utilizing scene information predicted from the corresponding timestamp in other frames. In addition to selecting adjacent frames $j \in \mathcal{N}(i)$, we also choose an extra frame $q$ from the global time to ensure the global consistency of the DCT trajectory.}
    \label{fig:cross-time}
\end{figure}

\noindent\textbf{Cross-time rendering.}
If we only rely on Eq.~\ref{eq:dy_rgb_loss} for reconstruction, the model may inadequately capture the temporal consistency in dynamic scenes, potentially leading to overfitting to input images and failing to produce high-quality novel views. To mitigate this issue, we utilize DCT trajectories to establish consistency between the $i$-th frame and other frames, as shown in Fig.~\ref{fig:cross-time}. Establishing temporal consistency between neighboring frames plays a critical role in the dynamic radiance field due to their strong correlation~\cite{gao2021dynamic, li2023dynibar}. However, blurry images relate to multiple images within the exposure time rather than a single image. Therefore, we express temporal consistency by establishing relationships between corresponding timestamps across different exposure times. Specifically, for a timestamp $t_l^i$ in an exposure time from frame $i$, the scene flow of a spatial point $\mathbf{x}$ from $t_l^i$ to timestamp $t_l^j$ can be represented as: 
\begin{equation}
    f_{\mathbf{x}}^{i,l}(j) = \mathcal{T}_{\mathbf{x}}^{t_l^i}(t_l^j) - \mathcal{T}_{\mathbf{x}}^{t_l^i}(t_l^i)\, ,
    \label{eq:scene flow}
\end{equation}
where $j \in \mathcal{N}(i)$ denotes the neighbor frames of $i$. The corresponding $3$D point of $\mathbf{x}$ at $t_l^j$ can be computed as $\mathbf{x}_l^{i \rightarrow j} = \mathbf{x} + f_{\mathbf{x}}^{i,l}(j)$. Establishing such relationships for every point on a ray $\mathbf{r}$ can obtain a warped ray $\mathbf{r}_l^{i \rightarrow j}$ from $t_l^i$ to $t_l^j$. Subsequently, we utilize volume rendering to obtain the color value of the warped ray $\mathbf{C}_l^{j \rightarrow i}(\mathbf{r})$: 
\begin{equation}
    \label{eq:warped}
    \mathbf{C}_l^{j \rightarrow i}(\mathbf{r}) = \int_{s_n}^{s_f} T_{t_l^j}(s) \thinspace \sigma_{t_l^j} (\mathbf{r}_l^{i \rightarrow j}(s)) \thinspace \mathbf{c}_{t_l^j}(\mathbf{r}_l^{i \rightarrow j}(s), \mathbf{d}_{t_l^i}) ds\, .
\end{equation}
Then we average the images within the exposure time to yield the blurred image $\mathbf{B}^{j \rightarrow i}(\mathbf{r})$, and we calculate the cross-time rendering loss between $\mathbf{B}^{j \rightarrow i}(\mathbf{r})$ and $\mathbf{B}_{\textrm{gt}}^i(\mathbf{r})$: 
\begin{equation}
    \mathbf{B}^{j \rightarrow i}(\mathbf{r}) = \frac{1}{n} \sum_{l=1}^{n} \mathbf{C}_l^{j \rightarrow i}(\mathbf{r})\, ,
\end{equation}
\begin{equation}
    \mathcal{L}_{\textrm{cross}} = \sum_{j \in \mathcal{N}(i)} \mathbf{W}^{j \rightarrow i}(\mathbf{r}) \Vert \mathbf{B}^{j \rightarrow i}(\mathbf{r}) - \mathbf{B}^{\textrm{gt}}_i(\mathbf{r}) \Vert_2^2 \, ,
\end{equation}
where $\mathbf{W}^{j \rightarrow i}(\mathbf{r})$ represents a motion disocclusion weight, which is the average outcome of $n$ disocclusion weights $\mathbf{W}_l^{j \rightarrow i}(\mathbf{r})$ computed by $\mathcal{W}_t$ in Eq.~\ref{eq:MLP}. For the specifics of disocclusion weight settings, we offer detailed information in the supplementary material. Through the aforementioned cross-time rendering, the DCT coefficients $\Psi$ predicted for the same point at different times remain consistent, which preserves the temporal consistency of the scene. 

In addition to using adjacent frames $j \in \mathcal{N}(i)$ for cross-time rendering, we include an additional frame $q$ for ensuring global time consistency. During the early training stages, $q$ is selected to be close to $i$. As training progresses, $q$ is gradually chosen further away from $i$ until it encompasses the entire global time range. We employ a similar rendering process to obtain the blurry image $\mathbf{B}^{q \rightarrow i}(\mathbf{r})$ and compute cross-time rendering loss with input image. In summary, the cross-time rendering loss can be written as:
\begin{equation}
    \mathcal{L}_{\textrm{cross}} = \sum_{j \in \{\mathcal{N}(i), q\}} \mathbf{W}^{j \rightarrow i}(\mathbf{r}) \Vert \mathbf{B}^{j \rightarrow i}(\mathbf{r}) - \mathbf{B}^{\textrm{gt}}_i(\mathbf{r}) \Vert_2^2 \, .
\end{equation}

\subsection{Training \& Optimization}
\label{sec:method_4}

\begin{figure}[t]
  \centering
   \includegraphics[width=1.0\linewidth]{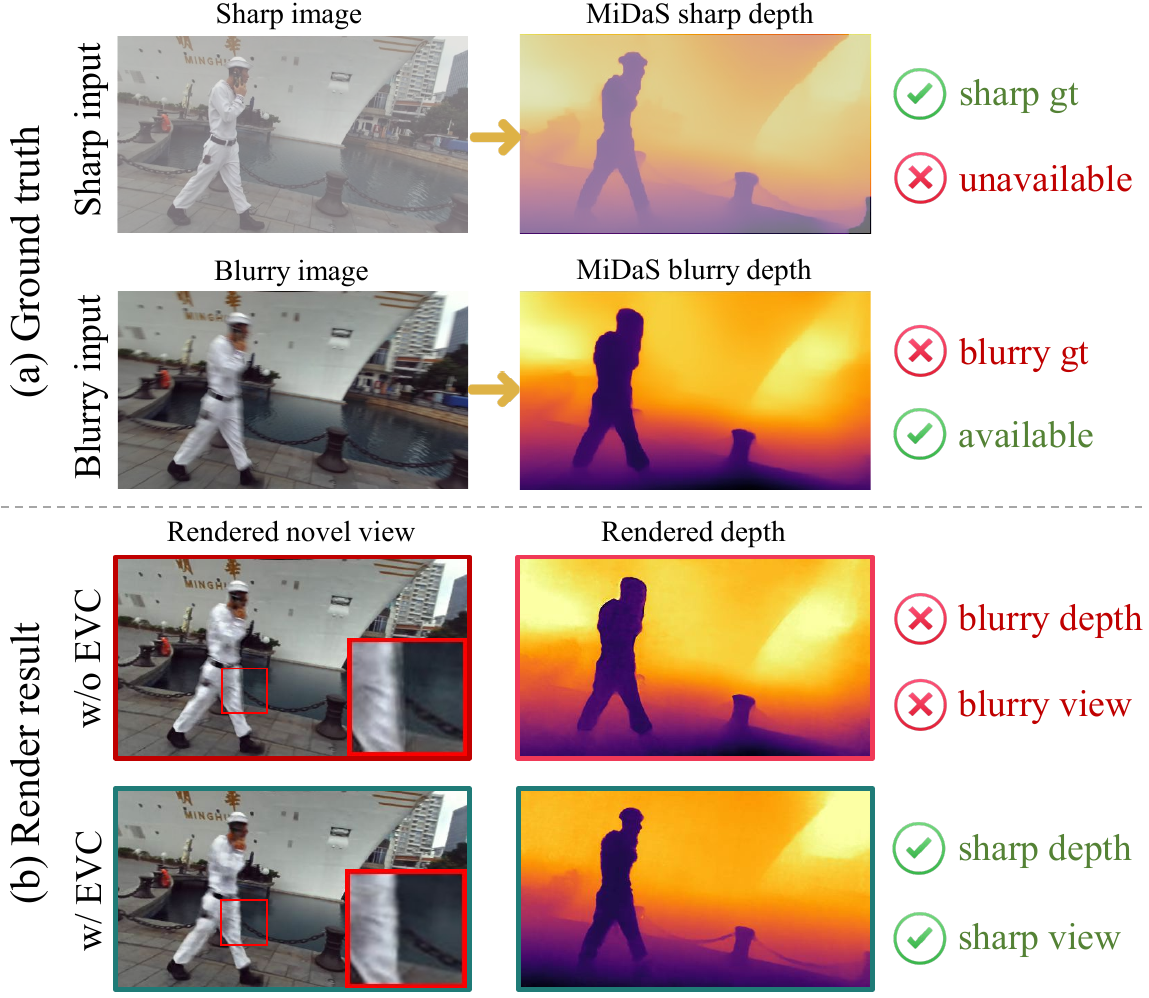}
   \caption{(a) We utilize depth map from MiDaS~\cite{ranftl2020towards} as the supervising signal for optimizing the model. Compared with sharp inputs, blurry images are available but lead to inaccurate depth ground truth. In the example, MiDaS interprets the blurry edges of the person as foreground, causing the person on the depth map to appear `fatter' than in the ideal sharp one. (b) Using the MiDaS blurry depth to optimize the model directly may predict inaccurate depth maps and distorted novel views. To mitigate this issue, we employ EVC to enhance our model for predicting sharp depth maps and novel views from blurry images.}
   \label{fig:data-driven}
\end{figure}

\noindent\textbf{Data-driven priors.}
We introduce data-driven priors $\mathcal{L}_{\textrm{data}}$ to constrain the scene geometry. We utilize RAFT~\cite{teed2020raft} and MiDaS~\cite{ranftl2020towards} to obtain ground truth optical flow and depth for the input blurry images. Intuitively, we only need to calculate the depth maps for each timestamp, and optical flow maps from each timestamp to the next exposure time, subsequently averaging them within a single exposure time and constraining them with the ground truth. However, we encountered issues where the depth and optical flow do not have a translucent effect in the blurry edges, as opposed to observed in RGB data (we show a depth map as an example in Fig.~\ref{fig:data-driven}(a)). Therefore, using simple averaging is unsuitable for constraining depth and optical flow. To address this issue, we implement Extreme Value Constraints (EVC) for both depth and optical flow maps. Specifically, after calculating the depth maps $\{\mathbf{D}_1, \mathbf{D}_2, \cdots, \mathbf{D}_n\}$ and optical flow maps $\{\mathbf{F}_1, \mathbf{F}_2, \cdots, \mathbf{F}_n\}$ projected from $3$D scene flow within an exposure time, we derive the predicted blurry result by taking the minimum depth value and maximum optical flow value, and calculate the loss with ground truth: 
\begin{equation}
    \begin{aligned}
        \mathbf{D}(\mathbf{r}) &= \min \{\mathbf{D}_1(\mathbf{r}), \mathbf{D}_2(\mathbf{r}), \cdots, \mathbf{D}_n(\mathbf{r})\}\, , \\
        \mathbf{F}(\mathbf{r}) &= \max \{\mathbf{F}_1(\mathbf{r}), \mathbf{F}_2(\mathbf{r}), \cdots, \mathbf{F}_n(\mathbf{r}) \}\, ,
    \end{aligned}
\end{equation}
\begin{equation}
\label{eq:data}
    \mathcal{L}_{\textrm{data}} = \Vert \mathbf{D}(\mathbf{r}) - \mathbf{D}_{\textrm{gt}}(\mathbf{r}) \Vert_1 + \Vert \mathbf{F}(\mathbf{r}) - \mathbf{F}_{\textrm{gt}}(\mathbf{r}) \Vert_1 \, .
\end{equation}
After applying EVC for data-driven constraints, a sharper result can be generated for each timestamp (as shown in Fig.~\ref{fig:data-driven}(b)). Please refer to supplementary material for principles and technical details of EVC.

\noindent\textbf{Scene flow modeling.}
For a more accurate representation of scene motion, we use the predicted DCT coefficients $\Psi$ to calculate the scene flow and impose constraints using $\mathcal{L}_{\textrm{sf}}$. Similar to NSFF~\cite{li2021neural}, $\mathcal{L}_{\textrm{sf}}$ consists of three components: cycle consistency of the scene flow, spatial-temporal smoothness constraints, and L$1$ regularization to the scene flow.

\noindent\textbf{Learning DCT basis.}
We model the motion trajectories using Eq.~\ref{eq:DCT}; however, we discovered that employing a learnable DCT basis produced superior outcomes. To achieve this, we initialize the DCT basis using the expression $\cos \left( \frac{\pi}{2*nN} (2t + 1)k \right)$, and treat them as learnable embeddings, facilitating joint learning with the models throughout the training process.

\noindent\textbf{Implementation details.}
In experiments, we set $n=7$ and $K=6$. We employ the Adam optimizer~\cite{kingma2014adam} to jointly optimize the static and dynamic MLPs, camera parameters, and DCT basis. The learning rates are set to $5 \times 10^{-4}$ for MLPs, $1 \times 10^{-3}$ for camera parameters, and $1.25 \times 10^{-4}$ for DCT basis. We train each scene for $300k$ iterations, which takes approximately two days on a single NVIDIA RTX A$6000$ GPU.

\begin{figure*}
    \centering
    \includegraphics[width=1.0\linewidth]{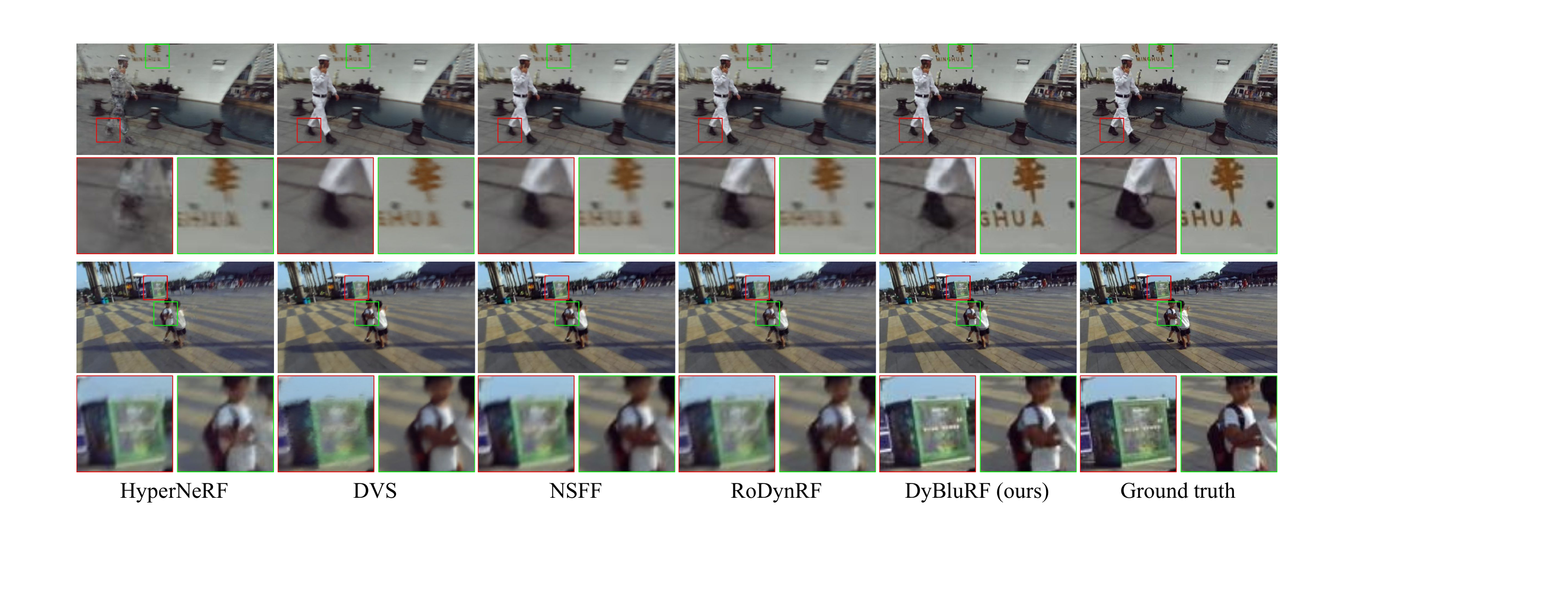}
    \caption{\textbf{Qualitative comparisons against all baselines.} Compared to existing dynamic NeRF methods, our method generates novel view images that are more faithful to the ground truth images, with less blur in both static and dynamic regions.}
    \label{fig:qualitative evaluation}
\end{figure*}
\section{Experiments}
\label{sec:exp}

\subsection{Datasets}
Since there is a lack of datasets specifically designed for addressing motion blur in dynamic neural radiance fields, we have curated a collection of dynamic scenes with motion blur from an existing Stereo Blur Dataset~\cite{zhou2019davanet}. 
This dataset was captured using a ZED stereo camera, containing sharp stereo image sequences along with their corresponding blurry image counterparts. Details about the generation of blur and other dataset information can be found in~\cite{zhou2019davanet}, and we also provide additional explanations in supplementary materials. To obtain accurate camera parameters for training neural radiance fields, we require image sequences of dynamic scenes with sufficient motion parallax. We collected $6$ dynamic scenes suitable for our task from this dataset. These scenes exhibit significant motion blur caused by both camera and object movements. 
In our experiments, we employ COLMAP~\cite{schonberger2016structure} to acquire the camera parameters for the input images.
For each scene, we extract $24$ frames from the original video, utilizing the left blurry image sequences for training and the corresponding right sharp image sequences for testing.

\begin{figure*}
    \centering
    \setlength{\tabcolsep}{0.7pt}
    \small
    \begin{tabular}{cccccc}
        \begin{minipage}[b]{0.343\columnwidth}
        \includegraphics[width=1\linewidth]{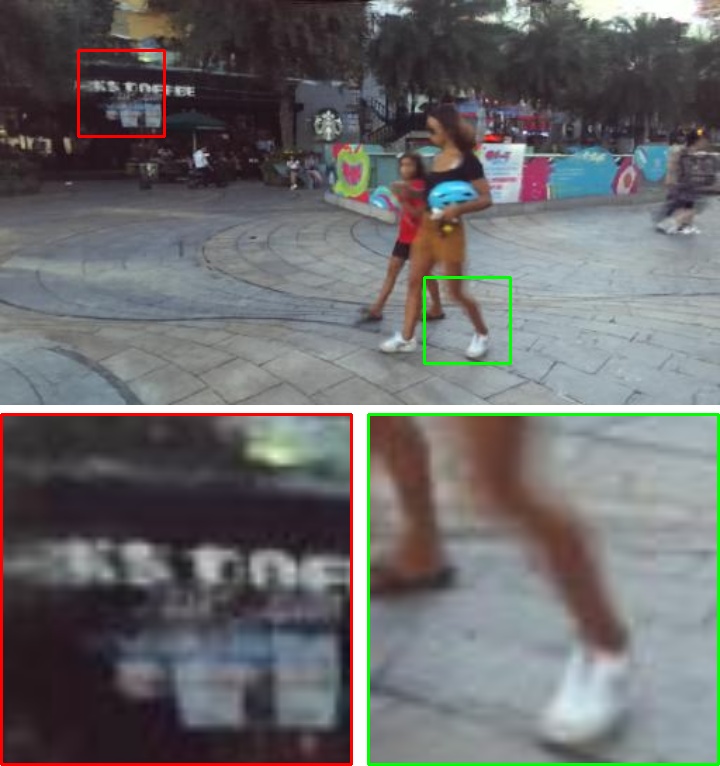}
        \end{minipage}
        &  
        \begin{minipage}[b]{0.343\columnwidth}
        \includegraphics[width=1\linewidth]{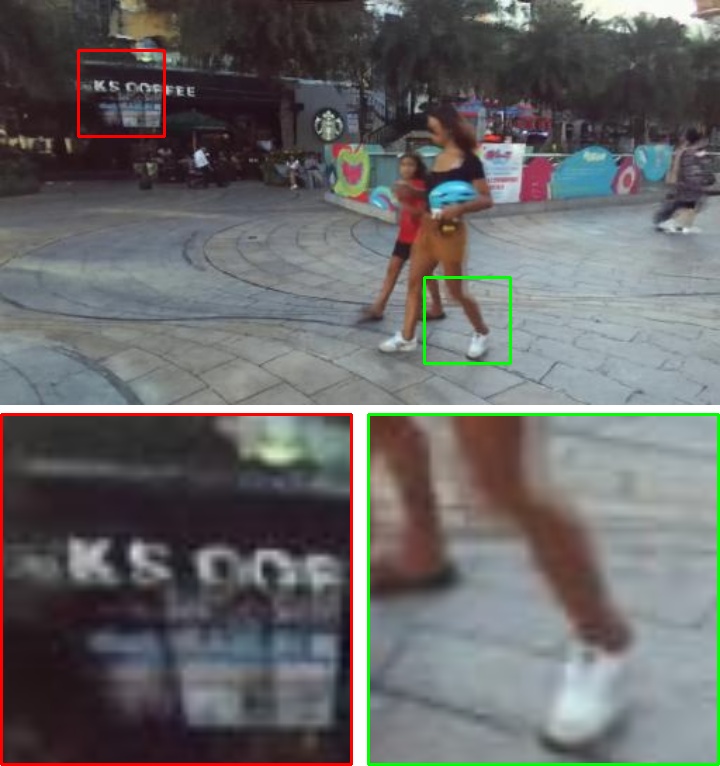}
        \end{minipage}
        &  
        \begin{minipage}[b]{0.343\columnwidth}
        \includegraphics[width=1\linewidth]{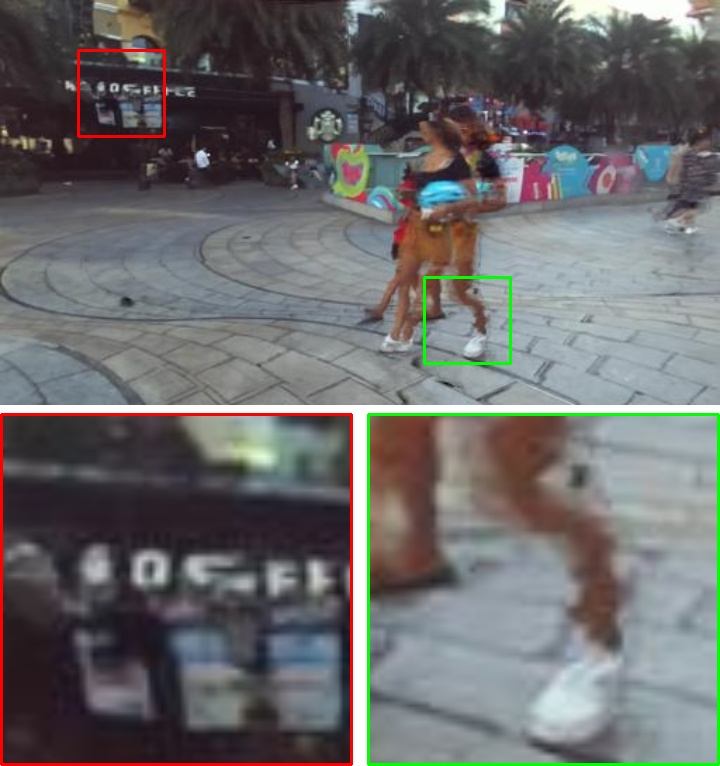}
        \end{minipage}
        &  
        \begin{minipage}[b]{0.343\columnwidth}
        \includegraphics[width=1\linewidth]{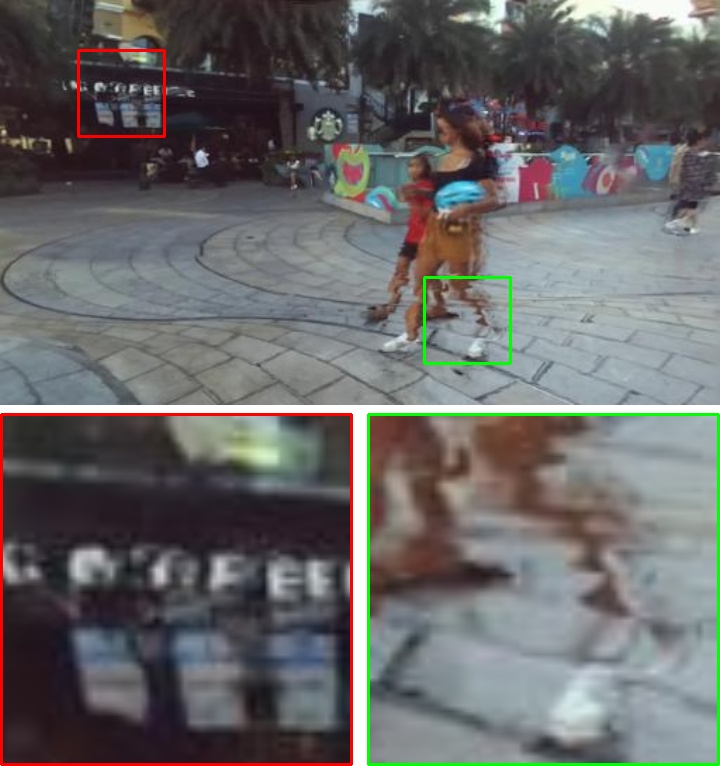}
        \end{minipage}
        &  
        \begin{minipage}[b]{0.343\columnwidth}
        \includegraphics[width=1\linewidth]{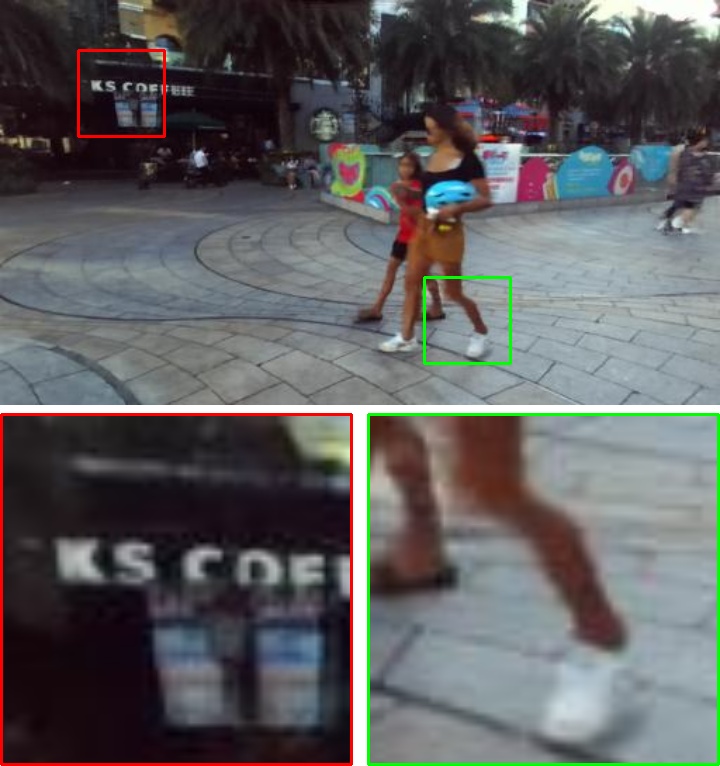}
        \end{minipage}
        &  
        \begin{minipage}[b]{0.343\columnwidth}
        \includegraphics[width=1\linewidth]{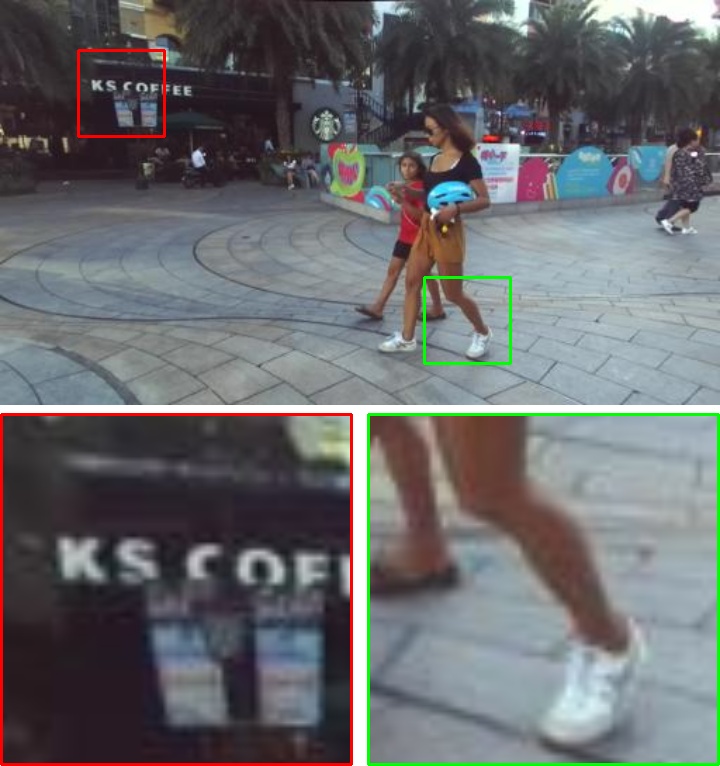}
        \end{minipage}
        \\
        \begin{minipage}[b]{0.343\columnwidth}
        \includegraphics[width=1\linewidth]{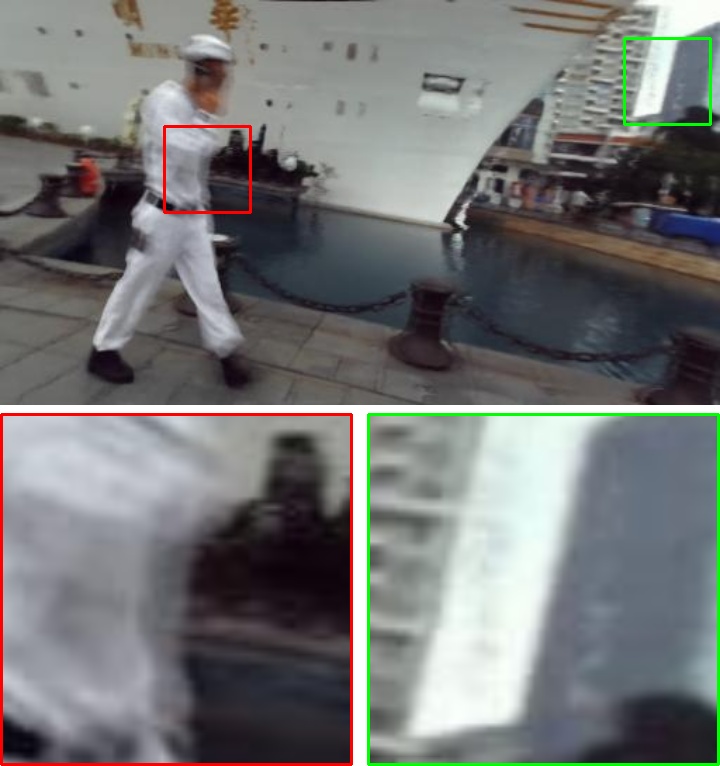}
        \end{minipage}
        &  
        \begin{minipage}[b]{0.343\columnwidth}
        \includegraphics[width=1\linewidth]{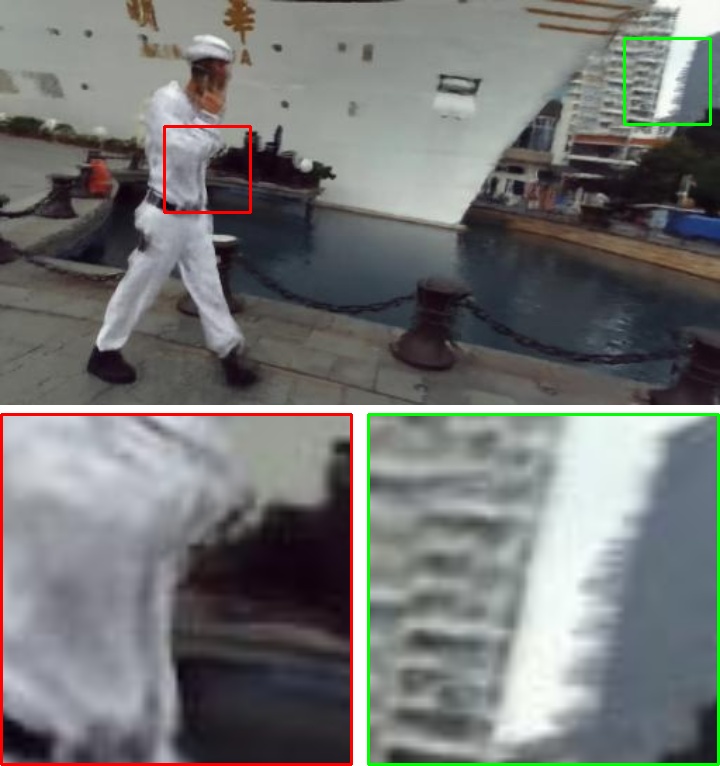}
        \end{minipage}
        &  
        \begin{minipage}[b]{0.343\columnwidth}
        \includegraphics[width=1\linewidth]{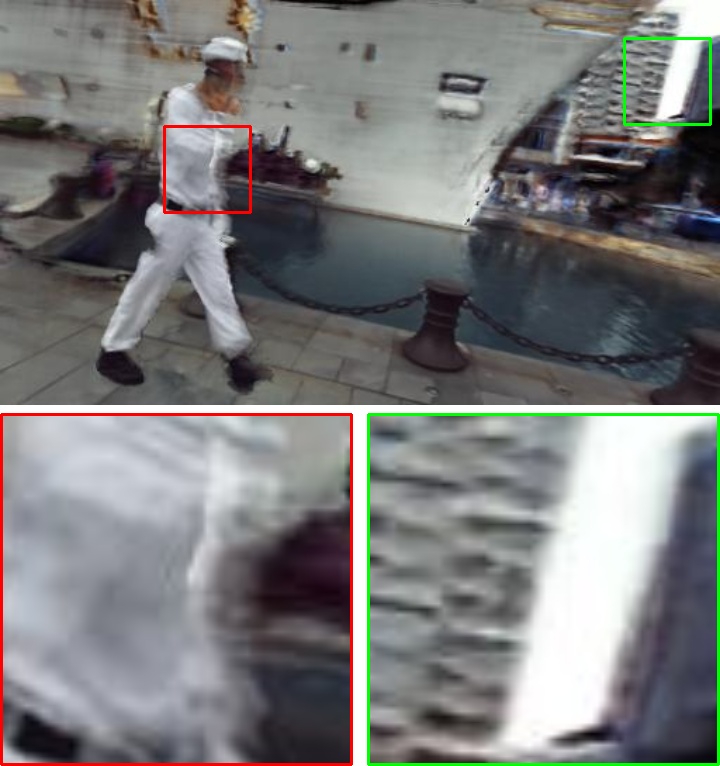}
        \end{minipage}
        &  
        \begin{minipage}[b]{0.343\columnwidth}
        \includegraphics[width=1\linewidth]{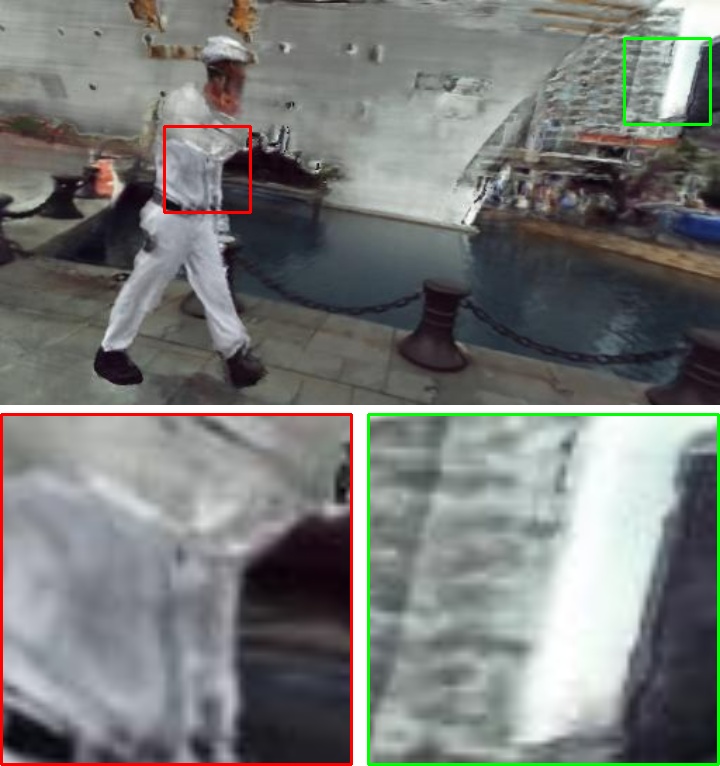}
        \end{minipage}
        &  
        \begin{minipage}[b]{0.343\columnwidth}
        \includegraphics[width=1\linewidth]{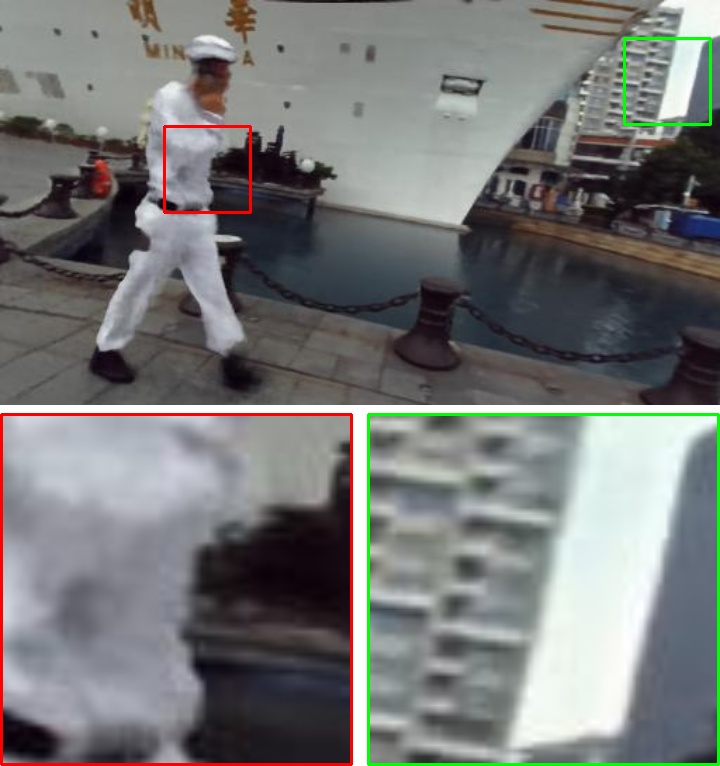}
        \end{minipage}
        &  
        \begin{minipage}[b]{0.343\columnwidth}
        \includegraphics[width=1\linewidth]{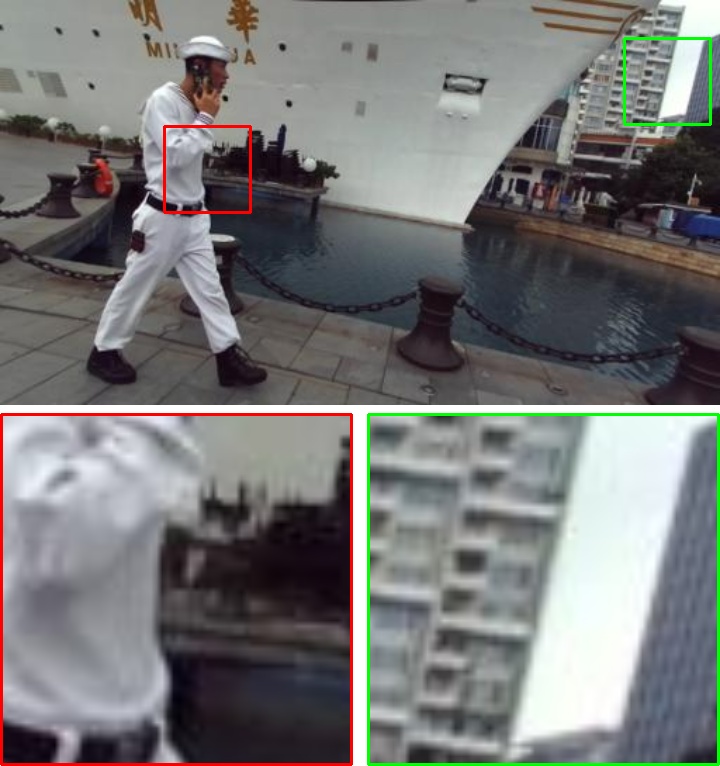}
        \end{minipage}
        \\ 
        \cite{zamir2022restormer} + \cite{li2021neural} & \cite{zhong2021towards} + \cite{li2021neural} & \cite{zamir2022restormer} + \cite{liu2023robust} & \cite{zhong2021towards} + \cite{liu2023robust} & DyBluRF (ours) & Ground truth \\
    \end{tabular}
    \vspace{-5pt}
    \caption{\textbf{Qualitative comparisons against dynamic NeRF baselines incorporated with $\mathbf{2}$D deblur method.} Even if we use preprocessed input blurry images by $2$D deblur approaches to train existing dynamic NeRF methods, our method also generates more reliable novel views, with less blur in both static and dynamic regions.}
    \label{fig:compare2}
    \vspace{-10pt}
\end{figure*}

\begin{table}[]\footnotesize
    \centering
    \setlength{\tabcolsep}{1.1pt}
    \renewcommand\arraystretch{1.2}
    \subfloat{
    \begin{tabular}{@{}lccc@{}}
        \toprule
        Methods & \scriptsize PSNR$\uparrow$ & \scriptsize SSIM$\uparrow$ & \scriptsize LPIPS$\downarrow$ \\
        \midrule
        HyperNeRF~\cite{park2021hypernerf} & 20.15 & 0.744 & 0.221 \\
        DVS~\cite{gao2021dynamic} & 22.30 & 0.811 & 0.193 \\
        NSFF~\cite{li2021neural} & \underline{23.33} & \underline{0.834} & \underline{0.181} \\
        RoDynRF~\cite{liu2023robust} & 19.28 & 0.748 & 0.217 \\
        \midrule
        DyBluRF (ours) & \textbf{25.66} & \textbf{0.895} & \textbf{0.079} \\
        \bottomrule
    \end{tabular}}\hskip1.2pt
    \subfloat{
    \begin{tabular}{@{}lccc@{}}
        \toprule
        Methods & \scriptsize PSNR$\uparrow$ & \scriptsize SSIM$\uparrow$ & \scriptsize LPIPS$\downarrow$ \\
        \midrule
        \cite{zamir2022restormer} + \cite{li2021neural} & 23.17 & 0.839 & 0.143 \\
        \cite{zhong2021towards} + \cite{li2021neural} & \underline{23.79} & \underline{0.854} & \underline{0.125} \\
        \cite{zamir2022restormer} + \cite{liu2023robust} & 20.76 & 0.787 & 0.162 \\
        \cite{zhong2021towards} + \cite{liu2023robust}  & 20.88 & 0.794 & 0.151 \\
        \midrule
        DyBluRF & \textbf{25.66} & \textbf{0.895} & \textbf{0.079} \\
        \bottomrule
    \end{tabular}}
    \caption{\textbf{Quantitative comparisons against all baselines.} The best performance is \textbf{boldfaced}, and the second is \underline{underlined}.}
    \label{tab:quantitative evaluation}
    \vspace{-10pt}
\end{table}

\subsection{Baselines and Error Metrics}
We compare our method against several state-of-the-art monocular dynamic NeRF methods. Specifically, 
the comparison involves a canonical space-based method, HyperNeRF~\cite{park2021hypernerf}, two scene flow-based methods, NSFF~\cite{li2021neural} and DVS~\cite{gao2021dynamic}, and a robust dynamic NeRF method, RoDynRF~\cite{liu2023robust}.
Furthermore, to demonstrate the superiority of our approach over 2D image deblurring, we select the current state-of-the-art single-image deblurring method Restormer~\cite{zamir2022restormer} and a dynamic scene video deblurring method JCD~\cite{zhong2021towards} to perform deblurring preprocessing on the input blurry images. We then train NSFF and RoDynRF on the preprocessed images for comparison. 

Similar to most existing dynamic NeRF methods, we use PSNR, SSIM, and LPIPS~\cite{zhang2018unreasonable} to quantitatively evaluate the quality of novel views generated by different approaches.

\subsection{Comparisons}
We conduct a quantitative evaluation 
on 
the scenes collected from the Stereo Blur Dataset. The average results across all scenes are shown in Tab.~\ref{tab:quantitative evaluation} (left), 
clearly demonstrating the superior performance of our method compared to the existing dynamic NeRF methods. 
We also provide qualitative comparisons of novel view results generated by different methods, as shown in Fig.~\ref{fig:qualitative evaluation}. Notably, existing dynamic NeRF methods cannot handle motion blur inputs, resulting in low-quality novel views in both dynamic and static regions. In contrast, our method can 
effectively manage inputs with motion blur and produce sharp and high-quality novel views.
Furthermore, to demonstrate the superiority of our method in modeling motion blur in $3$D space, we first preprocess the input blurry images with $2$D deblurring and then train dynamic NeRF models, e.g., RoDynRF, for comparison. As shown in Tab.~\ref{tab:quantitative evaluation} (right) and Fig.~\ref{fig:compare2}, our method continues to outperform the alternatives. This comparison underscores that $2$D deblurring methods fail to preserve the consistency of scene information. When the dynamic NeRFs are trained on these pre-processed images, it leads to undesirable artifacts in scene geometry. In contrast, our method models the blurring process in $3$D space, ensuring spatial-temporal consistency of scene geometry. 

\begin{figure}[t]
  \centering
   \includegraphics[width=1.0\linewidth]{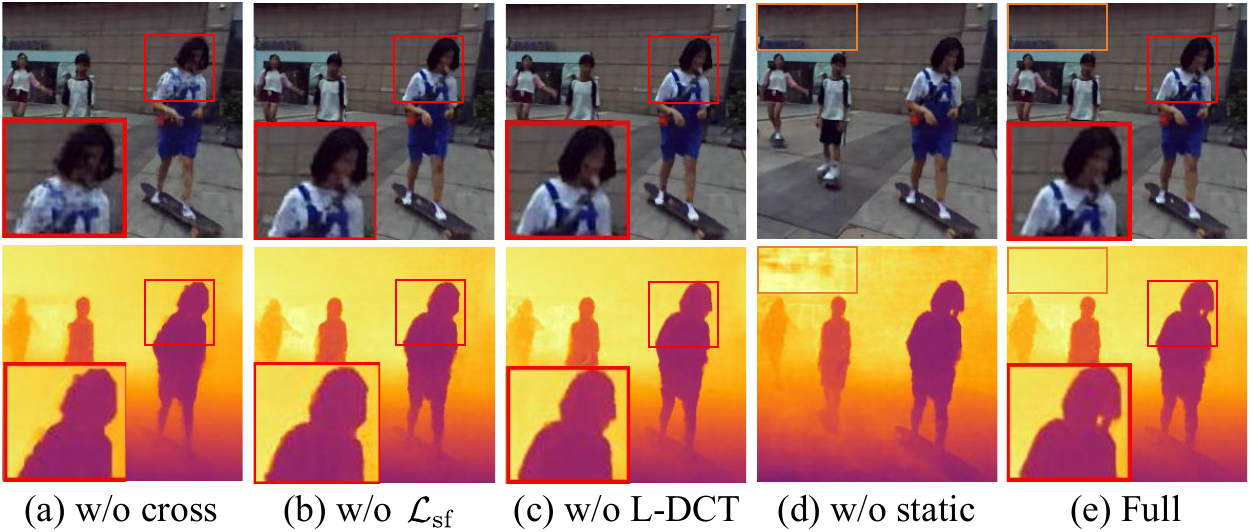}
    \vspace{-15pt}
   \caption{\textbf{Qualitative ablations.} (a) No cross-time rendering or (b) no $\mathcal{L}_{\textrm{sf}}$ results in degradation of predicted RGB and depth maps, especially containing artifacts in dynamic regions. (c) Not learning DCT basis may impact the representation of scene details. (d) The absence of static model leads to poor results in static regions. (e) Our proposed full method yields high-quality results in both dynamic and static regions.}
   \label{fig:ablation}
   \vspace{-13pt}
\end{figure}

\subsection{Ablation Study}
We conduct an ablation study to evaluate the contribution of each component in our model. Specifically, we evaluate the impact of ($1$) removing the cross-time rendering (w/o cross), ($2$) using average in data-driven priors instead of EVC (w/o EVC), ($3$) removing the static model (w/o static), ($4$) removing $\mathcal{L}_{\textrm{sf}}$ (w/o $\mathcal{L}_{\textrm{sf}}$), and ($5$) directly computing DCT basis using cosine values without being set as learnable parameters (w/o L-DCT). We report quantitative results in Tab.~\ref{tab:ablation study}. One can see that without cross-time rendering and scene flow constraints, view synthesis quality degrades significantly. Combining static model can also improve performance. Furthermore, removing EVC or not learning DCT basis can reduce performance, demonstrating the value of these training tricks for improving novel view quality. We also conduct visual comparisons in Fig.~\ref{fig:data-driven}(b) and Fig.~\ref{fig:ablation}, demonstrating that each component contributes to the overall performance of the model. Specifically, without EVC in data-driven constraints, the incorrect depth and flow priors can misleadingly optimize the model to predict low-quality rendering results. Removing cross-time rendering or $\mathcal{L}_{\textrm{sf}}$ cannot ensure the temporal consistency of dynamic scenes, resulting in obvious artifacts in dynamic regions. Static model can improve the scene representation for static elements, benefiting high-quality rendering results in static scenes. Finally, learning DCT basis can improve the representation of scene details. 

\begin{table}[]
    \centering
    \small
    \setlength{\tabcolsep}{5.0pt}
    \renewcommand\arraystretch{0.8}
    \begin{tabular}{@{}lccc@{}}
        \toprule
        Methods & PSNR$\uparrow$ & SSIM$\uparrow$ & LPIPS$\downarrow$ \\
        \midrule
        w/o cross & 24.54 & 0.875 & 0.087 \\
        w/o EVC & 25.42 & 0.890 & \underline{0.079} \\
        w/o static & 24.67 & 0.869 & 0.090 \\
        w/o $\mathcal{L}_{\textrm{sf}}$ & 24.97 & 0.879 & 0.085 \\
        w/o L-DCT  & \underline{25.49} & \underline{0.891} & 0.085 \\
        \midrule
        Full & \textbf{25.66} & \textbf{0.895} & \textbf{0.079}\\
        \bottomrule
    \end{tabular}
    \caption{\textbf{Ablation study.} The best performance is \textbf{boldfaced}, and the second is \underline{underlined}.}
    \label{tab:ablation study}
    \vspace{-10pt}
\end{table}
\section{Conclusion}
\label{sec:conclusion}


We propose DyBluRF, a dynamic radiance field to synthesize sharp novel views from a monocular video with motion blur. By modeling camera trajectory and object motion trajectories within the scene, our method overcomes the impact of motion blur inputs on novel view synthesis. We also ensure the global temporal consistency of the predicted DCT trajectory through cross-time rendering. We conducted extensive experiments using scenes collected from the Stereo Blur Dataset, which demonstrate that our method outperforms existing dynamic NeRF methods in handling monocular inputs with motion blur.



{
    \small
    \bibliographystyle{ieeenat_fullname}
    \bibliography{main}
}

\clearpage
\setcounter{page}{1}
\maketitlesupplementary


\section{Appendix}
We provide the following content in this supplementary:
\begin{itemize}
    \item Detailed description of disocclusion weights; 
    \item Principles and technical details of EVC; 
    \item Detailed description of scene flow constrains $\mathcal{L}_{\textrm{sf}}$;
    \item Final loss;
    \item More information on the dataset;
    \item Detailed quantitative and qualitative results;
    \item Results on sharp inputs;
    \item Limitations.
\end{itemize}

\renewcommand\thesubsection{\Alph{subsection}}

\subsection{Disocclusion weights}
Recall in Eq.~\ref{eq:MLP} of our main paper, the dynamic MLP additionally outputs a disocclusion weight $\mathcal{W}_t = (w_{\textrm{fw}}, w_{\textrm{bw}})$. The weights $w_{\textrm{fw}}$ and $w_{\textrm{bw}} \in [0, 1]$ represent the probability of occlusion for a spatial point. They determine the confidence regarding occlusion occurring from a specific timestamp in the current frame to the corresponding timestamp in adjacent frames. Certainly, a value closer to $0$ signifies minimal chances of the point being occluded during that interval. Conversely, a value closer to $1$ indicates a higher likelihood of the point experiencing occlusion or disocclusion. Specifically, for a timestamp $t_l^i$ from frame $i$, where $l \in \{1, \cdots, n\}$ represents the specific timestamp in an exposure time, the disocclusion weights are denoted as $\mathcal{W}_{t_l^i} = (w_l^{i \rightarrow {i + 1}}, w_l^{i \rightarrow {i - 1}})$. To obtain the disocclusion weight map in $2$D plane, the weight along the ray $\mathbf{r}_l^i$ is used for volume rendering with opacity from adjacent frames: 
\begin{equation}
    \label{eq:disocclusion}
    \mathbf{W}_l^{j \rightarrow i}(\mathbf{r}_l^i) = \int_{s_n}^{s_f} T_{t_l^j}(s) \thinspace \sigma_{t_l^j} (\mathbf{r}_l^{i \rightarrow j}(s)) \thinspace (1 - w_l^{i \rightarrow j}(\mathbf{r}_l^i(s)) ds\, ,
\end{equation}
where $j \in \mathcal{N}(i) = \{i+1, i-1\}$ denotes the adjacent frames of frame $i$, $\mathbf{r}_l^{i \rightarrow j}$ represents the warped ray mentioned by Eq.~\ref{eq:warped} in the main paper. 

For each timestamp within the exposure time, the disocclusion weight map $\mathbf{W}_l^{j \rightarrow i}(\mathbf{r})$ is computed using Eq.~\ref{eq:disocclusion}. Subsequently, we average these maps to yield the motion disocclusion weight $\mathbf{W}^{j \rightarrow i}(\mathbf{r})$ for cross-time rendering: 
\begin{equation}
    \mathbf{W}^{j \rightarrow i}(\mathbf{r}) = \frac{1}{n} \sum_{l=1}^n \mathbf{W}_l^{j \rightarrow i}(\mathbf{r})\, .
\end{equation}

\subsection{Extreme Value Constraints (EVC)}

\begin{figure}[t]
  \centering
   \includegraphics[width=1.0\linewidth]{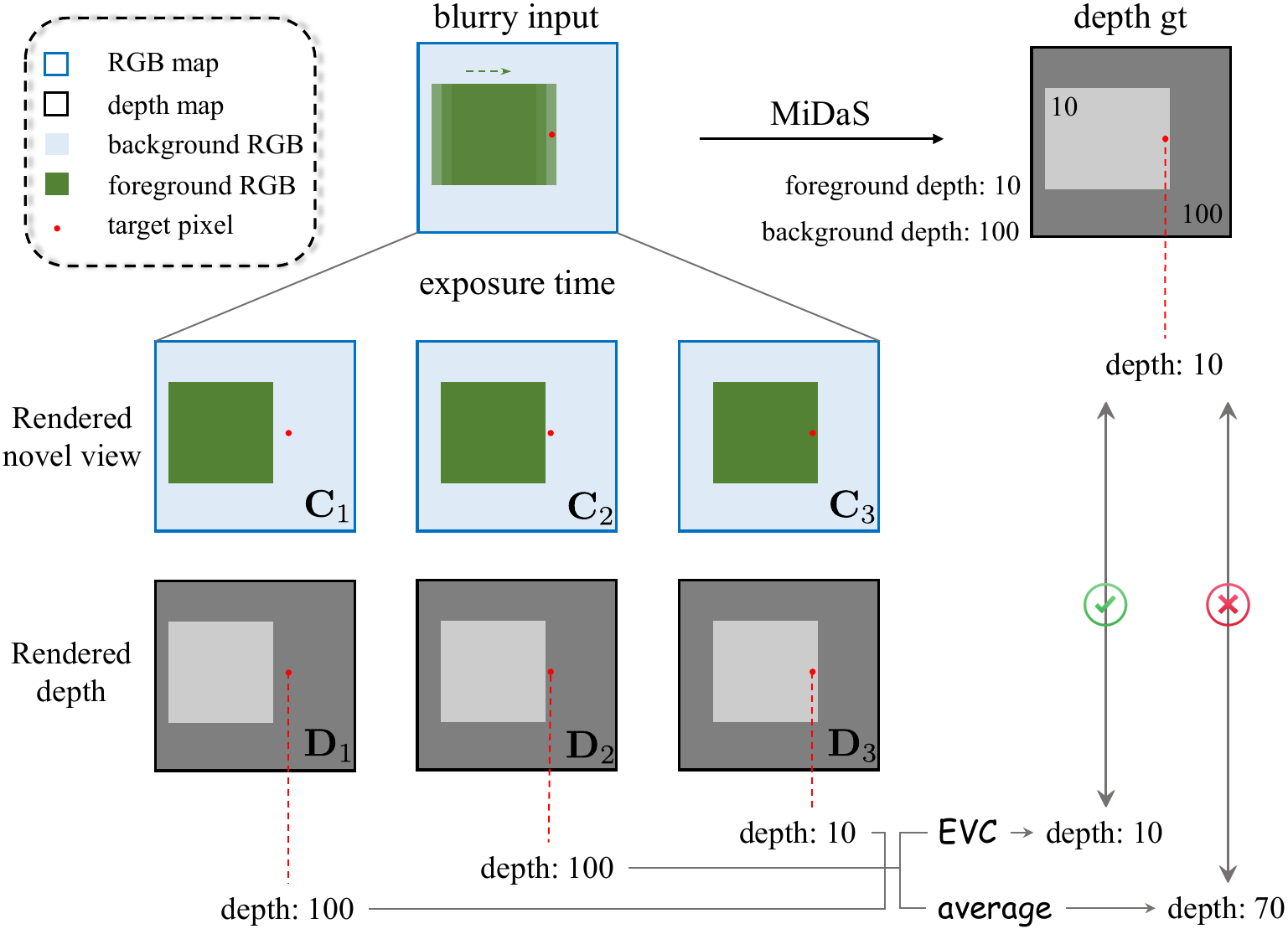}
   \caption{\textbf{Principle of EVC.}}
   \label{fig:EVC_supp}
   \vspace{-17pt}
\end{figure}

In this section, we will introduce the operational principles and technical details of EVC. In Fig.~\ref{fig:data-driven} of our main paper, we highlight an issue encountered while predicting depth and optical flow from blurry inputs. The depth prediction network and optical flow prediction network often mistakenly interpret the blurry edges in the RGB image as foreground, resulting in inaccuracies in data priors. To enable model learning accurate scene geometry by using these inaccurate data priors, we propose the EVC data prior constraint method. 

Fig.~\ref{fig:EVC_supp} provides an illustration of a simulation experiment on the working principle of EVC. We suppose that each exposure time is represented by three sharp images (\ie, $n=3$), with only one green square foreground moving from left to right in the scene, while the rest is a blue background. Given an input blurry frame, the model produces $3$ RGB images and depth maps within the exposure time. We consider a pixel position in blurry edges of the input frame, where the pixel changes the affiliation from the background (in $\textbf{C}_1$ and $\textbf{C}_2$) to the foreground (in $\textbf{C}_3$). To ease the discussion, we set the depth of the foreground to be $10$ while that of the background to be $100$ respectively. An intuitive practice to simulate the depth of input blurry frames may be averaging the three depth values to a value of $70$. However, due to the mistaken interpretation of blurry edges, MiDaS~\cite{ranftl2020towards} identifies the red point as the foreground, leading to its predicted depth value of $10$. Therefore the intuitive constraint cannot work for our method, manifesting as `foreground-fatter' depth map predictions as shown in the first row of Fig.~\ref{fig:data-driven}(b) in the main paper. In order to accommodate the depth prediction results, we should recognize a position as on the foreground (the green square), once it is covered by the foreground in a certain timestamp during the exposure time. This can be expressed as choosing the minimum depth value at the red-pointed position among the three sharp images (as EVC does).
This approach enables learning accurate scene geometry and rendering a sharp depth map, as shown in the second row of Fig.~\ref{fig:data-driven}(b) in the main paper. 

The process of data prior constraint for optical flow closely resembles that of depth. The model generates an optical flow map for each timestamp. This map represents the scene motion from the timestamp of current exposure time to the corresponding timestamp of adjacent exposure time. However, in contrast to depth where foreground values are typically smaller than background ones, optical flow for the foreground often exceeds that of the background. According to the principle and process of EVC, we use the maximum optical flow as the simulated blurry optical flow and compare it with the ground truth predicted by RAFT~\cite{teed2020raft}. In practice, we obtain the optical flow by projecting the predicted $3$D scene flow onto a $2$D plane.

\subsection{Scene flow modeling details}
Recall in Sec.~\ref{sec:method_4}, $\mathcal{L}_{\textrm{sf}}$ is used as a regularization loss for the scene flow calculated by Eq.~\ref{eq:scene flow} in the main paper. $\mathcal{L}_{\textrm{sf}}$ consists of three components: 
scene flow cycle consistency, spatial-temporal smoothness, and minimal scene flow. 

Scene flow cycle consistency enforces the forward scene flow $f_{\mathbf{x}}^{i,l}(j)$ of sampled $3$D points at timestamps $t_l^i$ consistent with the backward scene flow $f_{\mathbf{x}}^{j,l}(i)$ at the corresponding location at timestamps $t_l^j$: 
\begin{equation}
    \mathcal{L}_{\textrm{cyc}} = \sum_{j \in \{i \pm 1\}} \sum_{l=1}^n (1 - w_l^{i \rightarrow j}) \Vert f_{\mathbf{x}}^{i,l}(j) + f_{\mathbf{x}}^{j,l}(i) \Vert_1\, ,
\end{equation}
where $w_l^{i \rightarrow j}$ denotes the disocclusion weights used to reduce consideration of occlusion points. 

The spatial-temporal smoothness is designed to maintain continuity in scene flow both spatially and temporally. To achieve spatial smoothness, we encourage consistency between the scene flows of adjacent spatial points. Specifically, we compute L$1$ loss between scene flows sampled at two neighboring spatial points along the ray $\textbf{r}_l^i$. For temporal smoothness, we minimize the sum of forward and backward scene flow for each sampled spatial point to ensure the smoothness of predicted DCT trajectories. The spatial-temporal smoothness can be expressed as: 
\begin{equation}
    \begin{aligned}
        \mathcal{L}_{\textrm{smooth}} &= \sum_{\mathbf{y} \in \mathcal{N}(\mathbf{x})} \sum_{j \in \{i \pm 1\}} \sum_{l=1}^n \Vert f_{\mathbf{x}}^{i,l}(j) - f_{\mathbf{y}}^{i,l}(j) \Vert_1 \\
        &+ \frac{1}{2} \sum_{l=1}^n \Vert f_{\mathbf{x}}^{i,l}(i+1) + f_{\mathbf{x}}^{i,l}(i-1) \Vert_2^2\, ,
    \end{aligned}
\end{equation}
where the first term denotes spatial smoothness and the second term denotes temporal smoothness. $\mathcal{N}(\mathbf{x})$ represents the neighboring spatial points of point $\mathbf{x}$ along the ray $\textbf{r}_l^i$. 

Finally, due to minor scene changes between adjacent frames, we additionally apply a minimal scene flow constraint to minimize the predicted $3$D scene flow: 
\begin{equation}
    \mathcal{L}_{\textrm{min}} = \sum_{j \in \{i \pm 1\}} \sum_{l=1}^n \Vert f_{\mathbf{x}}^{i,l}(j) \Vert_1\, .
\end{equation}

\begin{table*}[]
    \centering
    \footnotesize
    \setlength{\tabcolsep}{1.4pt}
    \renewcommand\arraystretch{1.2}
    \begin{tabular}{@{}lcccccccccccccccccc@{}}
    \toprule
        \multirow{2}{*}{Methods} & \multicolumn{3}{c}{Sailor} & \multicolumn{3}{c}{Seesaw} & \multicolumn{3}{c}{Street} & \multicolumn{3}{c}{Children} & \multicolumn{3}{c}{Skating} & \multicolumn{3}{c}{Basketball} \\ 
        \cmidrule(r){2-4} \cmidrule(r){5-7} \cmidrule(r){8-10} \cmidrule(r){11-13} \cmidrule(r){14-16} \cmidrule(r){17-19}
        & \scriptsize PSNR$\uparrow$ & \scriptsize SSIM$\uparrow$ & \scriptsize LPIPS$\downarrow$ & \scriptsize PSNR$\uparrow$ & \scriptsize SSIM$\uparrow$ & \scriptsize LPIPS$\downarrow$ & \scriptsize PSNR$\uparrow$ & \scriptsize SSIM$\uparrow$ & \scriptsize LPIPS$\downarrow$ & \scriptsize PSNR$\uparrow$ & \scriptsize SSIM$\uparrow$ & \scriptsize LPIPS$\downarrow$ & \scriptsize PSNR$\uparrow$ & \scriptsize SSIM$\uparrow$ & \scriptsize LPIPS$\downarrow$ & \scriptsize PSNR$\uparrow$ & \scriptsize SSIM$\uparrow$ & \scriptsize LPIPS$\downarrow$ \\ 
        \midrule
        BAD-NeRF~\cite{wang2023bad} & 16.96 & 0.631 & 0.333 & 20.58 & 0.795 & 0.220 & 20.27 & 0.670 & 0.190 & 18.10 & 0.650 & 0.386 & 19.08 & 0.691 & 0.345 & 17.99 & 0.731 & 0.271 \\
        HyperNeRF~\cite{park2021hypernerf} & 18.56 & 0.743 & 0.275 & 20.25 & 0.779 & 0.182 & 19.99 & 0.662 & 0.137 & 21.36 & 0.762 & 0.279 & 19.52 & 0.702 & 0.319 & 21.21 & 0.818 & 0.136 \\
        DVS~\cite{gao2021dynamic} & \underline{22.32} & \underline{0.810} & \underline{0.247} & 18.14 & 0.775 & 0.158 & 19.06 & 0.669 & 0.176 & 24.00 & 0.823 & 0.307 & 26.18 & 0.907 & 0.124 & 24.10 & 0.880 & 0.143 \\
        NSFF~\cite{li2021neural} & 19.06 & 0.726 & 0.290 & 19.92 & 0.807 & 0.178 & \underline{23.42} & \underline{0.800} & \underline{0.121} & \underline{24.55} & \underline{0.846} & \underline{0.259} & \textbf{27.96} & \textbf{0.923} & \underline{0.116} & \underline{25.06} & \underline{0.903} & \underline{0.122} \\
        RoDynRF~\cite{liu2023robust} & 12.69 & 0.584 & 0.317 & \underline{23.71} & \textbf{0.894} & \underline{0.116} & 19.80 & 0.719 & 0.161 & 15.02 & 0.609 & 0.382 & 21.66 & 0.831 & 0.160 & 22.82 & 0.850 & 0.166 \\
        \midrule
        DyBluRF (ours) & \textbf{23.50} & \textbf{0.860} & \textbf{0.115} & \textbf{24.56} & \underline{0.882} & \textbf{0.075} & \textbf{26.88} & \textbf{0.906} & \textbf{0.068} & \textbf{25.57} & \textbf{0.884} & \textbf{0.092} & \underline{27.94} & \underline{0.917} & \textbf{0.072} & \textbf{25.28} & \textbf{0.920} & \textbf{0.050} \\
        \bottomrule
    \end{tabular}
    \caption{\textbf{Quantitative comparisons for every scene against all dynamic NeRF baselines.} The best performance is \textbf{boldfaced}, and the second is \underline{underlined}.}
    \label{tab:quantitative_individual_1}
\end{table*}

\begin{table*}[]
    \centering
    \footnotesize
    \setlength{\tabcolsep}{1.4pt}
    \renewcommand\arraystretch{1.2}
    \begin{tabular}{@{}lcccccccccccccccccc@{}}
    \toprule
        \multirow{2}{*}{Methods} & \multicolumn{3}{c}{Sailor} & \multicolumn{3}{c}{Seesaw} & \multicolumn{3}{c}{Street} & \multicolumn{3}{c}{Children} & \multicolumn{3}{c}{Skating} & \multicolumn{3}{c}{Basketball} \\ 
        \cmidrule(r){2-4} \cmidrule(r){5-7} \cmidrule(r){8-10} \cmidrule(r){11-13} \cmidrule(r){14-16} \cmidrule(r){17-19}
        & \scriptsize PSNR$\uparrow$ & \scriptsize SSIM$\uparrow$ & \scriptsize LPIPS$\downarrow$ & \scriptsize PSNR$\uparrow$ & \scriptsize SSIM$\uparrow$ & \scriptsize LPIPS$\downarrow$ & \scriptsize PSNR$\uparrow$ & \scriptsize SSIM$\uparrow$ & \scriptsize LPIPS$\downarrow$ & \scriptsize PSNR$\uparrow$ & \scriptsize SSIM$\uparrow$ & \scriptsize LPIPS$\downarrow$ & \scriptsize PSNR$\uparrow$ & \scriptsize SSIM$\uparrow$ & \scriptsize LPIPS$\downarrow$ & \scriptsize PSNR$\uparrow$ & \scriptsize SSIM$\uparrow$ & \scriptsize LPIPS$\downarrow$ \\ 
        \midrule
        \cite{zamir2022restormer} + \cite{li2021neural} & 18.96 & 0.728 & 0.252 & 20.19 & 0.827 & 0.149 & \underline{23.39} & 0.790 & 0.125 & \underline{24.88} & \underline{0.864} & \underline{0.182} & \underline{26.91} & \textbf{0.922} & \textbf{0.067} & \underline{24.70} & \underline{0.903} & 0.084 \\
        \cite{zhong2021towards} + \cite{li2021neural} & \underline{22.60} & \underline{0.805} & \underline{0.150} & 21.31 & 0.856 & 0.135 & 23.06 & \underline{0.791} & \underline{0.114} & 24.27 & 0.853 & 0.195 & 26.88 & \underline{0.917} & 0.078 & 24.64 & \underline{0.903} & \underline{0.076} \\
        \cite{zamir2022restormer} + \cite{liu2023robust} & 17.16 & 0.703 & 0.227 & 23.63 & \underline{0.892} & 0.090 & 19.75 & 0.727 & 0.149 & 17.41 & 0.665 & 0.282 & 23.22 & 0.869 & 0.110 & 23.38 & 0.868 & 0.114 \\
        \cite{zhong2021towards} + \cite{liu2023robust} & 17.28 & 0.772 & 0.225 & \underline{23.85} & \textbf{0.893} & \underline{0.080} & 19.69 & 0.700 & 0.153 & 18.39 & 0.683 & 0.226 & 22.54 & 0.849 & 0.126 & 23.51 & 0.867 & 0.098 \\
        \midrule
        DyBluRF (ours) & \textbf{23.50} & \textbf{0.860} & \textbf{0.115} & \textbf{24.56} & 0.882 & \textbf{0.075} & \textbf{26.88} & \textbf{0.906} & \textbf{0.068} & \textbf{25.57} & \textbf{0.884} & \textbf{0.092} & \textbf{27.94} & \underline{0.917} & \underline{0.072} & \textbf{25.28} & \textbf{0.920} & \textbf{0.050} \\
        \bottomrule
    \end{tabular}
    \caption{\textbf{Quantitative comparisons for every scene against dynamic NeRF methods with blurry image preprocess.} The best performance is \textbf{boldfaced}, and the second is \underline{underlined}.}
    \label{tab:quantitative_individual_2}
    \vspace{-7pt}
\end{table*}

\subsection{Final loss}
Considering the RGB image rendering, temporal consistency of dynamic scenes, data-driven constraints, and scene flow modeling, the final training loss of our method is:
\begin{equation}
    \mathcal{L} = \mathcal{L}_{\textrm{RGB}} + \mathcal{L}_{\textrm{cross}} + \lambda_{\textrm{data}} \mathcal{L}_{\textrm{data}} + \lambda_{\textrm{sf}} \mathcal{L}_{\textrm{sf}}\, ,
\end{equation}
where
\begin{equation}
    \begin{aligned}
        \mathcal{L}_{\textrm{RGB}} &= \mathcal{L}_{\textrm{RGB}}^{\textrm{cb}} + \mathcal{L}_{\textrm{RGB}}^{\textrm{dy}} + \lambda_{\textrm{st}} \mathcal{L}_{\textrm{RGB}}^{\textrm{st}}\, , \\
        \mathcal{L}_{\textrm{sf}} &= \lambda_{\textrm{cyc}} \mathcal{L}_{\textrm{cyc}} + \mathcal{L}_{\textrm{smooth}} + \mathcal{L}_{\textrm{min}}\, ,
    \end{aligned}
\end{equation}
$\lambda_{\textrm{data}}$, $\lambda_{\textrm{sf}}$, $\lambda_{\textrm{st}}$, $\lambda_{\textrm{cyc}}$ denote the weights for $\mathcal{L}_{\textrm{data}}$, $\mathcal{L}_{\textrm{sf}}$, $\mathcal{L}_{\textrm{RGB}}^{\textrm{st}}$ and $\mathcal{L}_{\textrm{cyc}}$, respectively. 
During training, $\lambda_{\textrm{data}}$ multiplies $0.1$ every $50k$ iterations to prevent the model from overfitting to data priors.

\subsection{Dataset}
We conduct experiments using dynamic scenes from the Stereo Blur Dataset~\cite{zhou2019davanet}. Similar to most datasets in image deblurring~\cite{nah2017deep, su2017deep}, the Stereo Blur Dataset generates motion blur by averaging a sharp high frame rate sequence. In practice, this dataset is captured using a ZED stereo camera capable of acquiring high frame rate ($60$fps) image sequences of dynamic scenes at a resolution of $1280 \times 720$. However, because the frame rate is still not high enough, direct averaging may lead to some unrealistic artifacts. Therefore, the dataset employs a video frame interpolation method~\cite{niklaus2017video} to increase the frame rate to $480$fps. Subsequently, averaging is performed on different numbers $(17, 33, 49)$ of consecutive frames to generate motion blur, with the sharp center frame among the consecutive frames serving as the ground truth for the blurry image. 

However, not all scenes in the Stereo Blur Dataset can be used in our work. That is because of two issues: $1)$ Many scenes within the dataset are static and lack moving objects. $2)$ Several scenes have minimal camera motion, resulting in a lack of motion blur caused by camera movement and insufficient motion parallax to obtain camera parameters. Therefore, we select $6$ dynamic scenes from the Stereo Blur Dataset that are tailored for NeRF-based methods. These scenes encompass both camera and object motion blur, showcasing varied size object movements like playing seesaw, walking, and skating. We employ COLMAP to acquire camera parameters from the image sequences and downsample the image resolution to $512 \times 288$ for experiments. Similar to NSFF, we process the input blurry image sequences through the MiDaS to obtain depth maps, use RAFT to generate optical flow maps, and employ an instance segmentation network (Mask r-cnn~\cite{he2017mask}) to derive motion masks for moving objects.

\subsection{Detailed results}
In this section, we present detailed quantitative and qualitative results of the comparative experiments in our main paper. In Tab.~\ref{tab:quantitative evaluation} of the main text, we have shown the average quantitative results of all baselines across the $6$ scenes. Here, we will provide individual quantitative results for each scene, as depicted in Tab.~\ref{tab:quantitative_individual_1} for blurry inputs and Tab.~\ref{tab:quantitative_individual_2} for deblurring preprocess inputs. We also conduct comparative experiments with BAD-NeRF, which is a deblurring NeRF method designed for static scenes, in Tab.~\ref{tab:quantitative_individual_1}. Due to its inability to represent dynamic scenes, the performance of BAD-NeRF is much lower than ours. We also include detailed qualitative comparison results, as illustrated in Fig.~\ref{fig:qualitative evaluation supp 1} and Fig.~\ref{fig:compare_supp2}. One can see that our method performs the best quantitative results in most scenes. Although our approach slightly trails NSFF in the Skating scene, the qualitative results of our method in the Skating scene are better than NSFF, regardless of whether blurry inputs or pre-processed sharp inputs, as shown in the second row of Fig.~\ref{fig:qualitative evaluation supp 1} and Fig.~\ref{fig:compare_supp2}. We speculate that slightly lower metrics in the Skating scene of our method could be attributed to the background in the Skating scene having an extensive low-texture area with less motion blur, which cannot fully showcase the superiority of our method in handling motion blur input through metrics. However, the qualitative results demonstrate that our method outperforms all baselines, especially in handling motion blur in dynamic scenes. 

From Tab.~\ref{tab:quantitative_individual_1} and Tab.~\ref{tab:quantitative_individual_2}, we can also observe that preprocessing blurry images with $2$D deblurring methods may yield inferior results compared to directly using blurry inputs. This phenomenon arises from the $2$D deblurring methods disrupting the temporal consistency of scene information, causing inaccurate scene representation in NeRF. This further reflects that our approach effectively ensures spatial-temporal consistency in dynamic scenes.

\begin{table*}[]
    \centering
    \footnotesize
    \setlength{\tabcolsep}{3.6pt}
    \renewcommand\arraystretch{1.0}
    \begin{tabular}{lcccccccccccc}
        \toprule
        \multirow{2}{*}{Methods} & \multicolumn{3}{c}{Balloon1} & \multicolumn{3}{c}{Balloon2} & \multicolumn{3}{c}{Jumping} & \multicolumn{3}{c}{Playground} \\
        \cmidrule(r){2-4} \cmidrule(r){5-7} \cmidrule(r){8-10} \cmidrule(r){11-13}
        & PSNR$\uparrow$ & SSIM$\uparrow$ & LPIPS$\downarrow$ & PSNR$\uparrow$ & SSIM$\uparrow$ & LPIPS$\downarrow$ & PSNR$\uparrow$ & SSIM$\uparrow$ & LPIPS$\downarrow$ & PSNR$\uparrow$ & SSIM$\uparrow$ & LPIPS$\downarrow$ \\
        \midrule
        NSFF~\cite{li2021neural} & \textbf{21.96} & \textbf{0.791} & 0.215 & 24.27 & 0.825 & 0.222 & 24.65 & 0.872 & 0.151 & 21.22 & 0.780 & 0.212 \\
        DyBluRF (ours) & 21.90 & 0.781 & \textbf{0.181} & \textbf{24.92} & \textbf{0.880} & \textbf{0.117} & \textbf{26.21} & \textbf{0.901} & \textbf{0.091} & \textbf{23.64} & \textbf{0.861} & \textbf{0.113} \\
        \midrule
        \midrule
        \multirow{2}{*}{Methods} & \multicolumn{3}{c}{Skating} & \multicolumn{3}{c}{Truck} & \multicolumn{3}{c}{Umbrella} & \multicolumn{3}{c}{Average} \\
        \cmidrule(r){2-4} \cmidrule(r){5-7} \cmidrule(r){8-10} \cmidrule(r){11-13}
        & PSNR$\uparrow$ & SSIM$\uparrow$ & LPIPS$\downarrow$ & PSNR$\uparrow$ & SSIM$\uparrow$ & LPIPS$\downarrow$ & PSNR$\uparrow$ & SSIM$\uparrow$ & LPIPS$\downarrow$ & PSNR$\uparrow$ & SSIM$\uparrow$ & LPIPS$\downarrow$ \\
        \midrule
        NSFF~\cite{li2021neural} & \textbf{29.29} & \textbf{0.936} & 0.129 & 25.96 & 0.863 & 0.167 & 22.97 & 0.769 & 0.295 & 24.33 & 0.834 & 0.199 \\
        DyBluRF (ours) & 28.36 & 0.913 & \textbf{0.087} & \textbf{30.01} & \textbf{0.946} & \textbf{0.043} & \textbf{24.19} & \textbf{0.831} & \textbf{0.163} & \textbf{25.60} & \textbf{0.873} & \textbf{0.114} \\
        \bottomrule
    \end{tabular}
    \caption{\textbf{Quantitative results on the non-blur dataset with NSFF.} The better performance is \textbf{boldface}.} 
    \label{tab:sharp_input}
\end{table*}


\subsection{Results on sharp inputs.}
Although our method is specifically tailored for dynamic scenes with motion blur, it remains capable of achieving comparable results to existing methods when dealing with sharp inputs. We evaluate our method on the Nvidia Dynamic Scene Dataset~\cite{yoon2020novel} without blur, configured with $n = 1$ for sharp image input. Here, we compare with NSFF, which performs second best in Tab.~\ref{tab:quantitative evaluation} in the main paper. As shown in Tab.~\ref{tab:sharp_input}, our method is marginally better than NSFF even in sharp inputs, which underscores our sustained efficacy even in sharp inputs. Meanwhile, our method significantly outperforms NSFF under blurry inputs in Tab.~\ref{tab:quantitative evaluation}. That demonstrates the effectiveness of our method in handling both motion blur and representing dynamic scenes. 

\begin{figure}[t]
  \centering
   \includegraphics[width=1.0\linewidth]{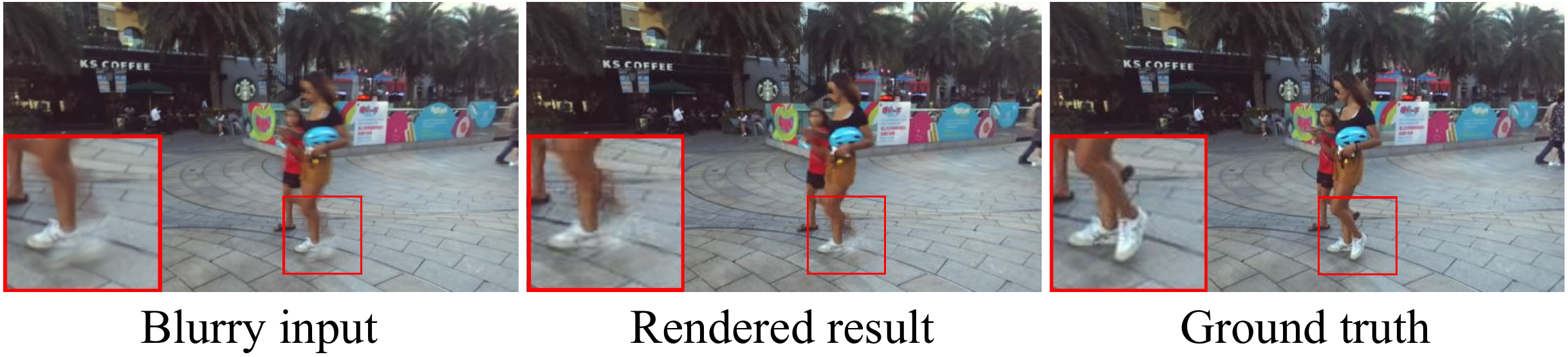}
   \caption{\textbf{Limitation.} It is challenging to restore a sharp image when input blur is caused by extreme object motion (\eg, the leg).}
   \label{fig:limitation}
\end{figure}

\subsection{Limitations}
Although our method can handle most of the motion blur in the input images, it might not be able to synthesize high-quality novel views when the motion blur is caused by highly complex and fast motion. As shown in Fig.~\ref{fig:limitation}, the rapid movement of the human leg causes extreme blurriness in the input image. This challenges our method in two aspects: firstly, excessive information loss of foreground may lead to the depth and flow prediction networks being unable to identify foreground objects, thereby affecting the representation of dynamic scenes. Secondly, the model tends to incorrectly learn the scene geometry supervised by such extremely blurry images, and gets stuck in the incorrect local minimum. In the future, we aim to explore combining explicit representations to enhance the temporal coherence of dynamic objects to solve this problem.

\begin{figure*}
    \centering
    \includegraphics[width=1.0\linewidth]{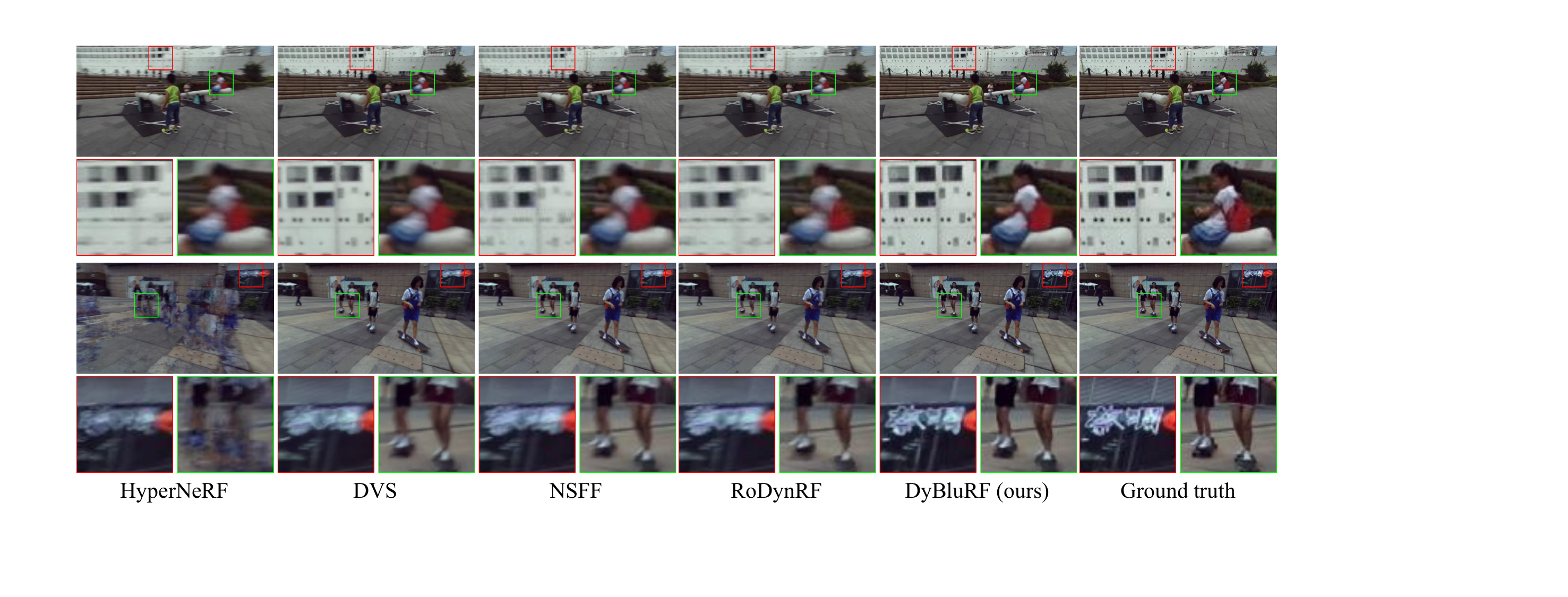}
    \caption{\textbf{Qualitative comparisons against all baselines.} Compared to existing dynamic NeRF methods, our method generates novel view images that are more faithful to the ground truth images, with less blur in both static and dynamic regions.}
    \label{fig:qualitative evaluation supp 1}
\end{figure*}

\begin{figure*}
    \centering
    \setlength{\tabcolsep}{0.7pt}
    \small
    \begin{tabular}{cccccc}
        \begin{minipage}[b]{0.343\columnwidth}
        \includegraphics[width=1\linewidth]{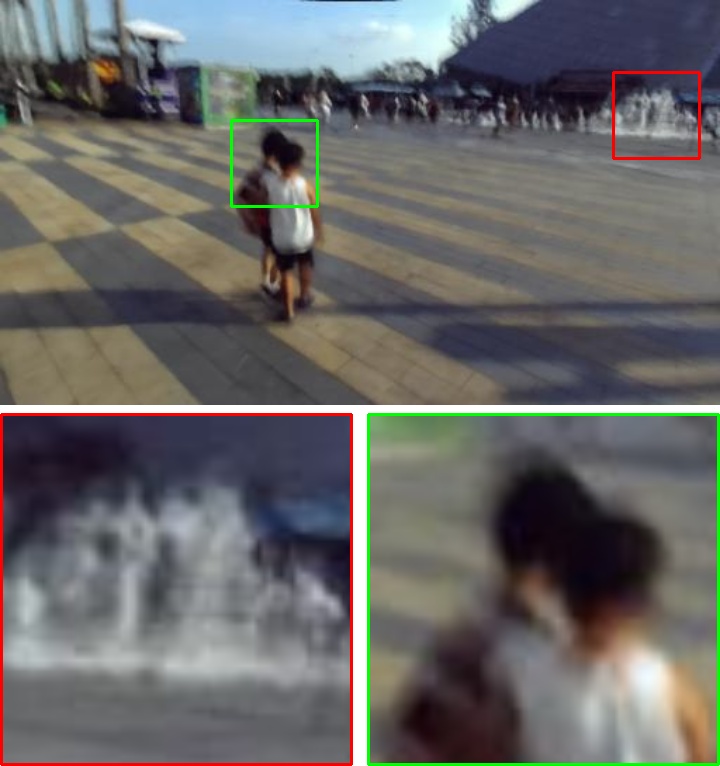}
        \end{minipage}
        &  
        \begin{minipage}[b]{0.343\columnwidth}
        \includegraphics[width=1\linewidth]{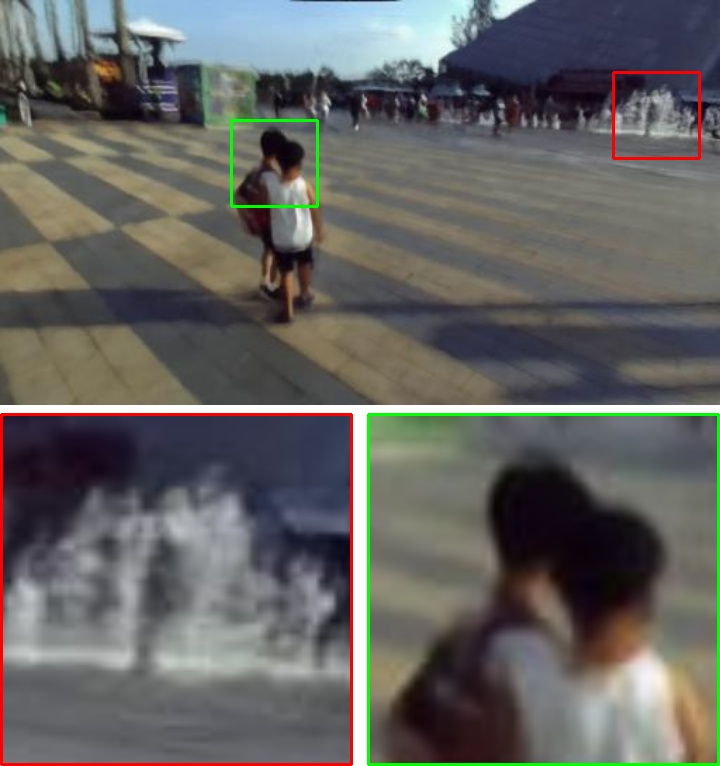}
        \end{minipage}
        &  
        \begin{minipage}[b]{0.343\columnwidth}
        \includegraphics[width=1\linewidth]{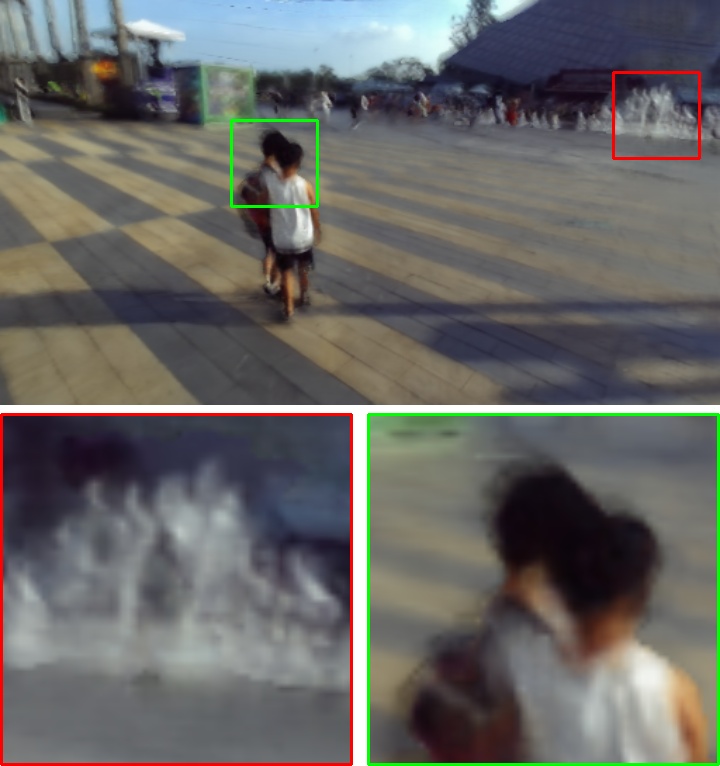}
        \end{minipage}
        &  
        \begin{minipage}[b]{0.343\columnwidth}
        \includegraphics[width=1\linewidth]{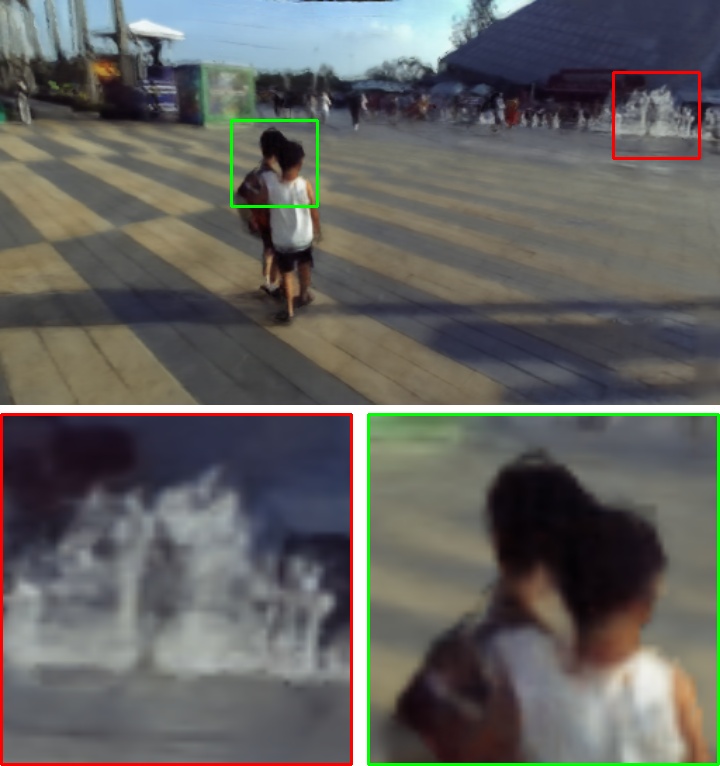}
        \end{minipage}
        &  
        \begin{minipage}[b]{0.343\columnwidth}
        \includegraphics[width=1\linewidth]{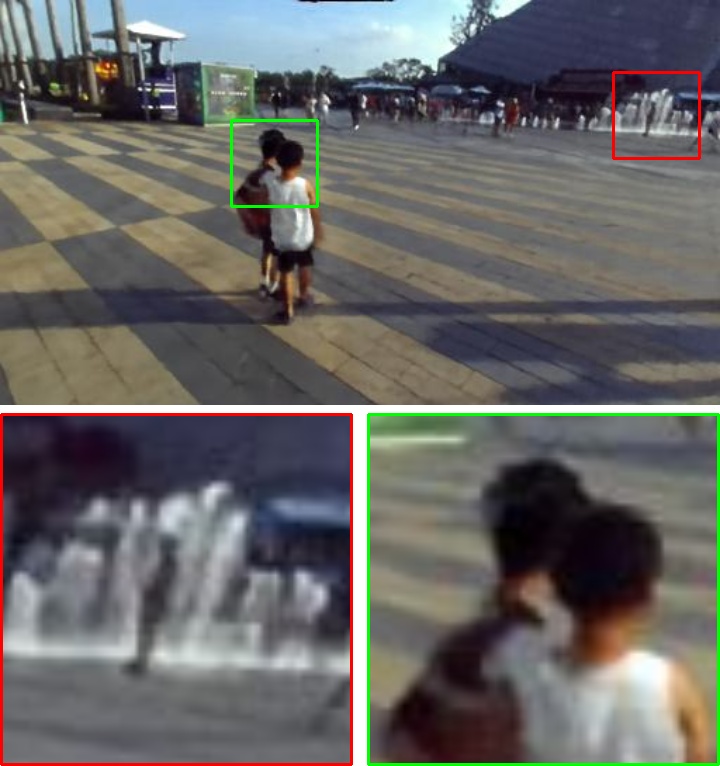}
        \end{minipage}
        &  
        \begin{minipage}[b]{0.343\columnwidth}
        \includegraphics[width=1\linewidth]{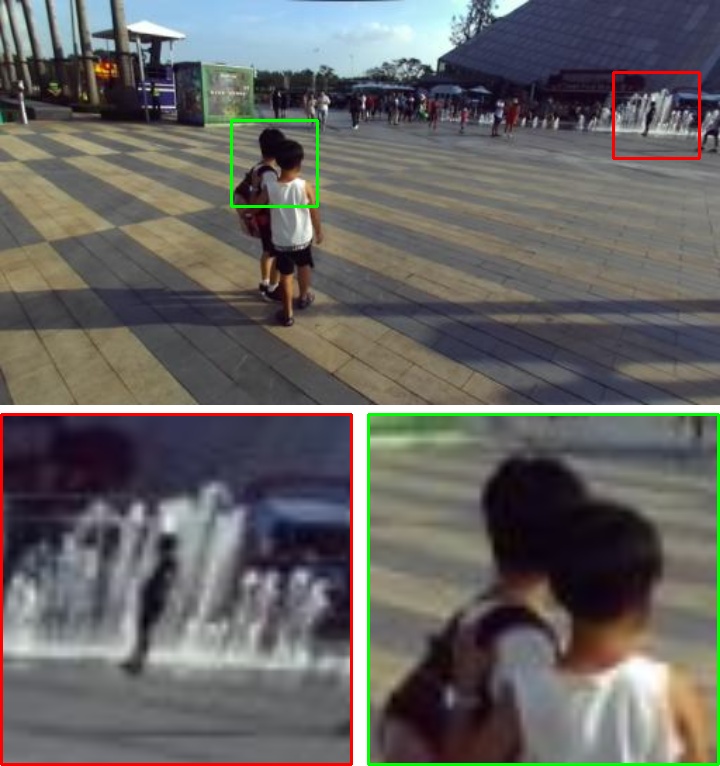}
        \end{minipage}
        \\
        \begin{minipage}[b]{0.343\columnwidth}
        \includegraphics[width=1\linewidth]{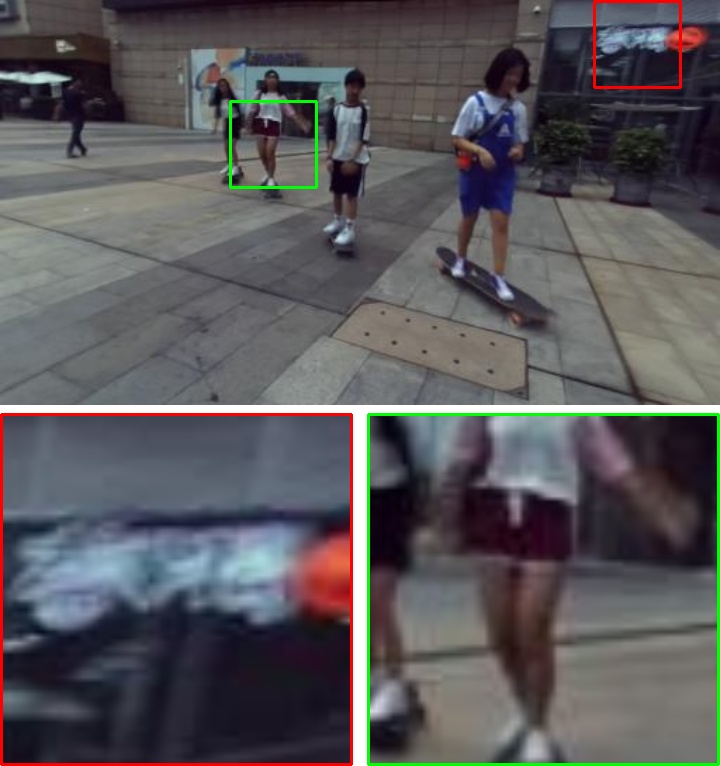}
        \end{minipage}
        &  
        \begin{minipage}[b]{0.343\columnwidth}
        \includegraphics[width=1\linewidth]{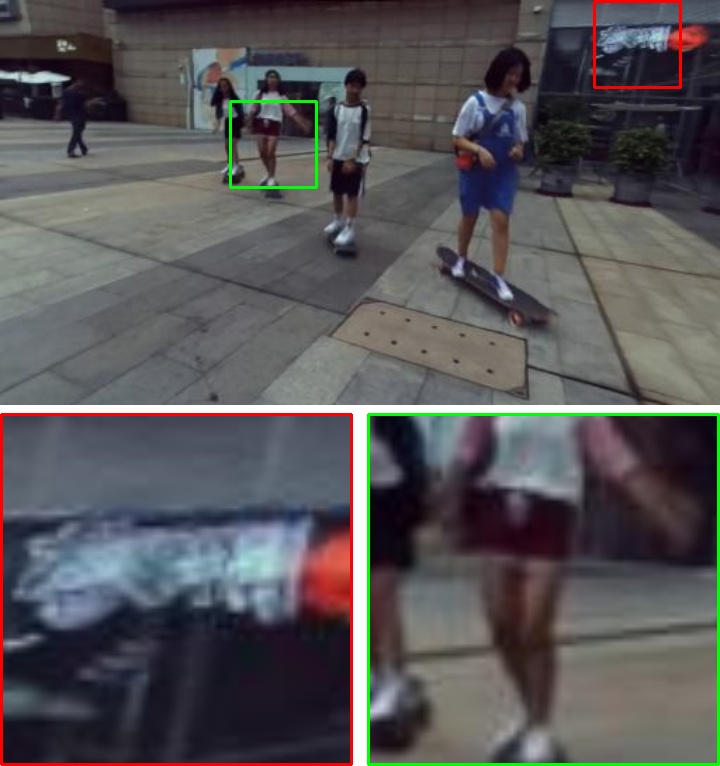}
        \end{minipage}
        &  
        \begin{minipage}[b]{0.343\columnwidth}
        \includegraphics[width=1\linewidth]{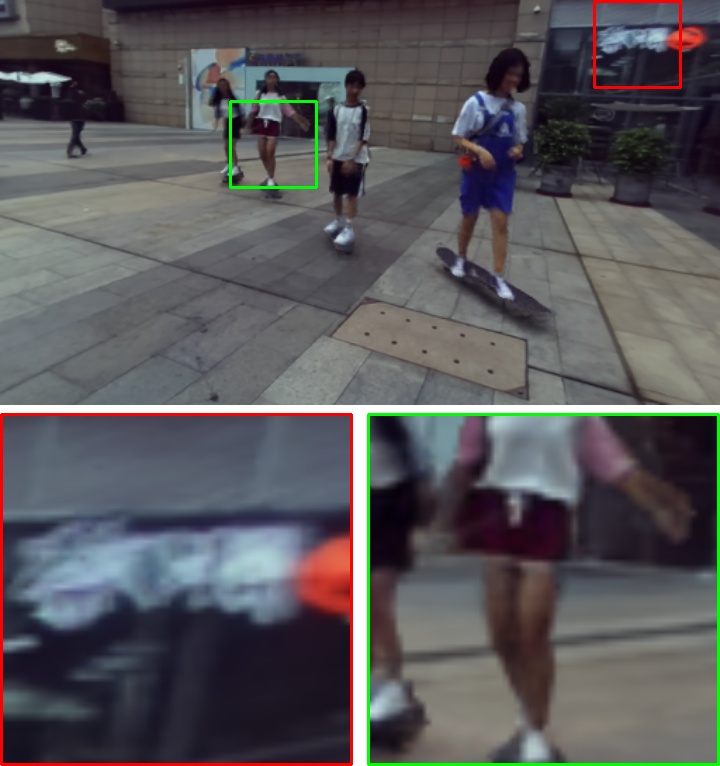}
        \end{minipage}
        &  
        \begin{minipage}[b]{0.343\columnwidth}
        \includegraphics[width=1\linewidth]{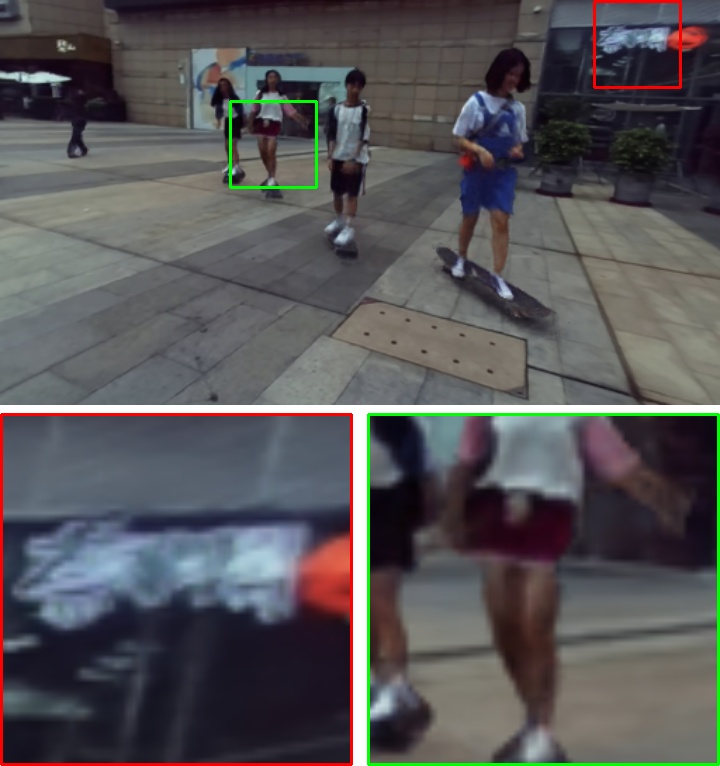}
        \end{minipage}
        &  
        \begin{minipage}[b]{0.343\columnwidth}
        \includegraphics[width=1\linewidth]{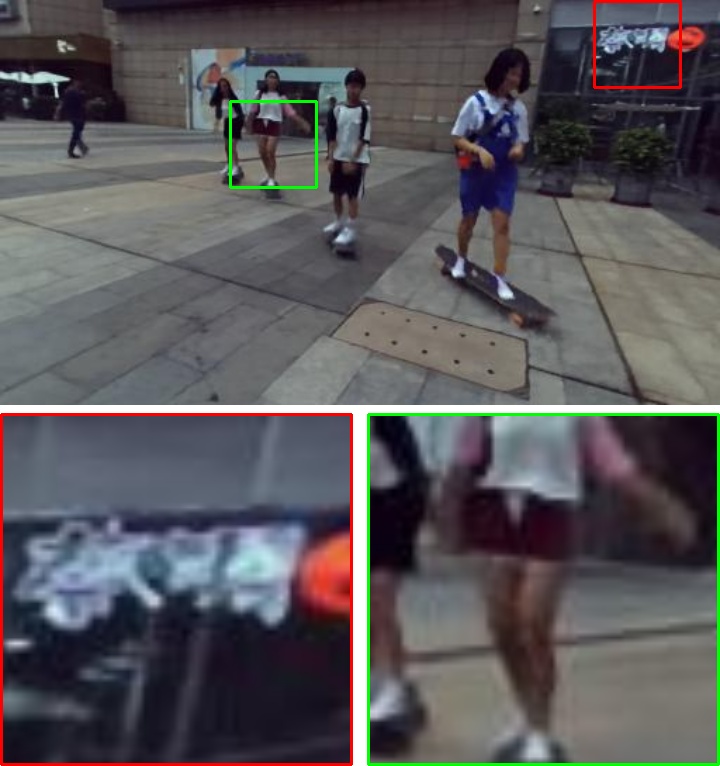}
        \end{minipage}
        &  
        \begin{minipage}[b]{0.343\columnwidth}
        \includegraphics[width=1\linewidth]{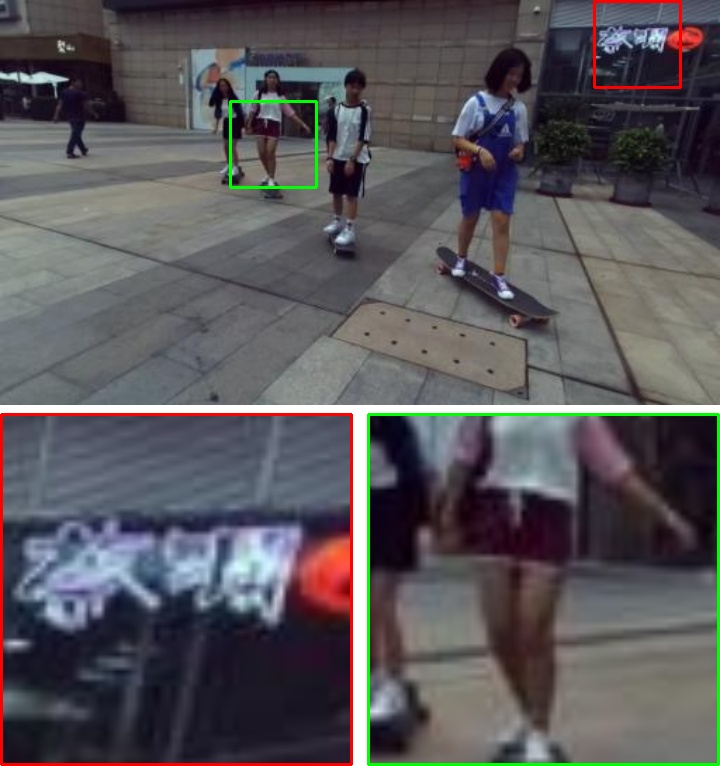}
        \end{minipage}
        \\ 
        \cite{zamir2022restormer} + \cite{li2021neural} & \cite{zhong2021towards} + \cite{li2021neural} & \cite{zamir2022restormer} + \cite{liu2023robust} & \cite{zhong2021towards} + \cite{liu2023robust} & DyBluRF (ours) & Ground truth \\
    \end{tabular}
    \vspace{-5pt}
    \caption{\textbf{Qualitative comparisons against dynamic NeRF baselines incorporated with $\mathbf{2}$D deblur method.} Even if we use preprocessed input blurry images by $2$D deblur approaches to train existing dynamic NeRF methods, our method also generates more reliable novel views, with less blur in both static and dynamic regions.}
    \label{fig:compare_supp2}
    \vspace{-10pt}
\end{figure*}


\end{document}